\documentclass[journal]{IEEEtran}
\usepackage{amsmath,amssymb,bm}
\usepackage{array,threeparttable,booktabs,multirow}
\usepackage{graphicx}
\usepackage[ruled]{algorithm2e}
\usepackage[noend]{algpseudocode}
\usepackage{cite}
\usepackage[dvipsnames]{xcolor}
\usepackage{setspace}
\usepackage[colorlinks=true, allcolors=blue]{hyperref}
\usepackage{pifont}
\usepackage{makecell}
\usepackage{diagbox}
\usepackage{setspace}
\usepackage{float}   
\usepackage{subfigure} 
\usepackage{enumitem}
\usepackage{colortbl}
\usepackage{amsthm}
\newtheorem{proposition}{Proposition}

\begin{document} 
\linespread{1.0}
\bstctlcite{BSTcontrol}

\title{
	\huge Latent Diffusion Model-Enabled Low-Latency Semantic Communication in the Presence of Semantic Ambiguities and Wireless Channel Noises
}
\author{
	Jianhua~Pei,~\IEEEmembership{Graduate~Student~Member,~IEEE}, %
	Cheng~Feng,~\IEEEmembership{Member,~IEEE}, %
	Ping~Wang,~\IEEEmembership{Fellow,~IEEE}, %
	Hina~Tabassum,~\IEEEmembership{Senior~Member,~IEEE}, %
	and~Dongyuan~Shi,~\IEEEmembership{Senior~Member,~IEEE}

	\thanks{
		Manuscript received July 8 2024; revised November 19 2024 and January 22 2025; accepted January 24 2025. This work was supported in part by the Natural Sciences and Engineering Research Council of Canada (NSERC) Discovery Grant funded by NSERC. The associate editor coordinating the review
		of this article and approving it for publication was G. Zhu. 
		\emph{(Corresponding author: Dongyuan Shi)}
	}%
	\thanks{
		Jianhua Pei and Dongyuan Shi are with the School of Electrical and Electronic Engineering, Huazhong University of Science and Technology, Wuhan, Hubei, China (e-mails: jianhuapei@hust.edu.cn;  dongyuanshi@hust.edu.cn). 
		
		Cheng Feng is with Energy Systems Engineering, System Engineering, Cornell University, Ithaca, NY, USA (e-mail: chengfeng@cornell.edu).  
		
		Ping Wang and Hina Tabassum are with the Department of Electrical Engineering and Computer Science, Lassonde School of Engineering, York University, Toronto, ON, Canada (e-mails: pingw@yorku.ca;  hinat@yorku.ca).
	}%
	
}
\maketitle

\begin{abstract}
	Deep learning (DL)-based Semantic Communications (SemCom) is becoming critical to maximize overall efficiency of communication networks. Nevertheless, SemCom is sensitive to wireless channel uncertainties, source outliers, and suffer from poor generalization bottlenecks. To address the mentioned challenges, this paper develops a latent diffusion model-enabled SemCom system with three key contributions, i.e., i) to handle potential outliers in the source data, semantic errors obtained by projected gradient descent based on the vulnerabilities of DL models, are utilized to update the parameters and obtain an outlier-robust encoder, ii) a lightweight single-layer latent space transformation adapter completes one-shot learning at the transmitter and is placed before the decoder at the receiver, enabling adaptation for out-of-distribution data and enhancing human-perceptual quality, and iii) an end-to-end consistency distillation (EECD) strategy is used to distill the diffusion models trained in latent space, enabling deterministic single or few-step low-latency denoising in various noisy channels while maintaining high semantic quality. Extensive numerical experiments across different datasets demonstrate the superiority of the proposed SemCom system, consistently proving its robustness to outliers, the capability to transmit data with unknown distributions, and the ability to perform real-time channel denoising tasks while preserving high human perceptual quality, outperforming the existing denoising approaches in semantic metrics such as multi-scale structural similarity index measure (MS-SSIM) and learned perceptual image path similarity (LPIPS).
\end{abstract}

\begin{IEEEkeywords}
	Semantic communication, latent diffusion model, GAN inversion, channel denoising, semantic ambiguity.
\end{IEEEkeywords}

\section{Introduction} 
\label{sec:introduction}
\IEEEPARstart{W}{ith} the booming development of artifical intelligence (AI), augmented and virtual reality \cite{siriwardhana2021survey}, 4K/6K straming \cite{huang2023virtual}, and the intelligent sensing devices for smart grids \cite{abrahamsen2021communication} and vehicles \cite{wu2022unmanned} within the internet of things (IoT), an efficient and reliable communication system becomes an essential component in the realm of 6-th generation (6G) communications \cite{chowdhury20206g}. In information and communication technology, joint source-channel coding (JSCC) \cite{8723589} is committed to the integrated design of source and channel codes for efficient transmission of data, leveraging Shannon information theory. However, classic JSCC techniques, employing coding methods for engineering applications such as JPEG \cite{wallace1992jpeg}, JPEG2000 \cite{christopoulos2000jpeg2000}, and BPG \cite{fan2018wide}, have solely focused on the statistical characteristics of the data being transmitted, disregarding the semantic content they encompass.

Recently, the pursuit of more efficient and intelligent feature extraction and data transmission has given rise to semantic communication (SemCom) systems \cite{yang2022semantic}, where the focus has shifted from traditional bit-level accuracy to the conveyance of meaning and intent. The essence of SemCom lies in its capacity to emphasize the transmission of semantic information, thus promising significant improvements in bandwidth utilization and overall communication efficiency \cite{luo2022semantic}. Fortunately, with the rapid advancement of machine learning, deep learning (DL)-based SemCom systems are becoming crucial \cite{xie2021deep}. Specifically, SemCom built upon neural networks such as variational autoencoder (VAE) \cite{dai2022nonlinear}, residual network (ResNet) \cite{he2016deep}, convolutional neural network (CNN) \cite{xu2021wireless}, long short-term memory (LSTM) network \cite{farsad2018deep}, generative adversarial network (GAN) \cite{9953099}, and Transformer \cite{xie2021deep} have demonstrated effectiveness in extracting the semantic features of source data. This allows for the mapping of source data into a lower-dimensional space for transmission over noisy wireless channels to the receiver, where it can ultimately be decoded back into its original form, whether that be images \cite{huang2022toward}, audio \cite{weng2021semantic}, text \cite{xie2021deep}, or multimodal data \cite{xie2022task}. \textbf{Nonetheless, the intrinsic complexity of semantic information, coupled with the uncertainity of communication channels, poses new challenges that some SemCom systems are not designed to handle. }

Diffusion models (DMs) have taken the forefront in the field of AI-generated content (AIGC) and have achieved remarkable advancements in generation quality \cite{ho2020denoising, song2020denoising}, surpassing other generative models in recent years. Consequently, the application of DMs to tackle challenges within SemCom systems is beginning to gain attraction \cite{du2023beyond, qiao2024latency}. Conditional DM, guided by semantic information from other users, progressively generate matching data for mixed reality applications \cite{du2023ai}. Similarly, conditional DMs, guided by invertible neural networks \cite{chen2024commin}, compressed one-hot maps \cite{grassucci2023generative}, decoded low-quality data \cite{yilmaz2023high}, and scene graphs \cite{yang2024sg2sc}, have been proposed for image transmission to achieve higher perceptual quality. DMs have also been adapted to rectify errors caused by channels with varying-fading gains and low signal-to-noise ratio (SNR) noises \cite{choukroun2022denoising}. Moreover, wireless channel estimation has also been performed by well-designed complex architectures based on DMs \cite{zilberstein2024joint, jiang2024large}. Besides serving as decoders for joint source-channel coding (JSCC) \cite{8723589}, DMs can also act as denoisers placed after decoders to enhance data quality \cite{jiang2024diffsc}. In \cite{grassucci2024diffusion}  and \cite{du2023exploring}, prompts, latent embeddings, or noisy data are transmitted over wireless channels to the receiver as starting points or input conditions for high-quality reverse process of DMs, inevitably increasing bandwidth burden. However, the primary bottleneck of DMs lies in their slow data generation speed due to the multi-step process in original high-dimensional pixel space required to improve reconstruction quality, making such time-consuming communication impractical for ultra low-latency SemCom and edge users. Thus, some denoising or encoding methods opt for latent DM (LDM) \cite{10480348} or acceleration techniques \cite{10104549, 10542391} to significantly reduce the computational complexity by only sampling in low-dimensional latent space. Nevertheless, since these enhanced approaches \cite{10480348, 10104549, 10542391} still feature a multi-step process during sampling, they inadequately address the challenges of real-time SemCom.

DMs and LDMs-based SemCom systems offer high perceptual quality but also introduces a bottleneck of high-latency. Moreover, structural errors, noises, and data following unknown distributions can introduce inaccuracies and distortions in the transmitted information when DL-based SemCom systems are deployed. The former, known as semantic errors, can arise from exploiting the vulnerabilities of DL models by adversarial attacks that lead to semantic discrepancies. Additionally, when a DL-based SemCom system trained by the specific category of data transmits out-of-distribution data \cite{9953099, zhang2022unified}, the reconstructed data at the receiver may also be semantically ambiguous due to the prevalent issues of poor generalization and overfitting in current DL models. To balance sampling quality and speed, LDM has been chosen as the underlying architecture for the SemCom approach due to its excellent semantic encoding, semantic decoding, and channel denoising capabilities \cite{rombach2022high}. Overall, the LDM-enabled SemCom also remains susceptible to semantic ambiguities from outliers or out-of-distribution data. Furthermore, when faced with noisy wireless channels, the LDM-enabled channel denoising method may not meet the low-latency SemCom requirements \cite{10480348}.

To address these issues, this paper presents a comprehensive framework that enhances low-latency SemCom by leveraging the capabilities of LDMs while simultaneously considering the effects of semantic ambiguities and channel imperfections. The proposed SemCom model builds upon and enhances the foundational architecture of a pretrained Wasserstein GAN \cite{adler2018banach} with VAE (VAE-WGAN). The overall contribution of this approach is threefold:

\begin{enumerate}[leftmargin=12pt]
	\item Semantic errors can significantly disrupt the normal semantic encoding and decoding of DL-based JSCC systems. To address this, the vulnerabilities of the pretrained encoder and generator are exploited using convex optimization to determine the most significant undetectable semantic errors. The pretrained encoder is then updated with the obtained semantic errors to refine the neural network parameters, making the encoder robust and resilient to anomalously transmitted data. This parameter update process with data augmentation is self-supervised.
	\item A rapid domain adaptation strategy is introduced to ensure the reconstructed data is semantically accurate at the receiver when the SemCom system transmits data with an unknown distribution. This strategy employs two additional lightweight single-layer neural networks that perform online one-shot or few-shot learning based on adversarial learning strategies. The updated parameters are transmitted to the dynamic neural network deployed at the receiver through the shared knowledge \cite{yang2022semantic} of the SemCom system, while the parameters of other networks remain unchanged, thus achieving low-cost latent space transformation.
	\item Inspired by channel denoising DM (CDDM) \cite{10480348} and consistency distillation \cite{song2023consistency}, the LDM based on ordinary differential equation (ODE) trajectories and variance explosion strategy is trained with known channel state information (CSI) \cite{zilberstein2024joint, jiang2024large}. During the sampling phase, it can denoise the received equalized signals according to different CSIs. Furthermore, the end-to-end consistency distillation (EECD) approach that considers semantic metrics is proposed to distill the trained LDM, ultimately transforming the multi-step denoising process into a deterministic one-step real-time denoising procedure, capable of flexibly addressing varying fading channels and uncertain SNRs.
\end{enumerate}
The efficiency and reliability of the proposed SemCom system in term of perceptual quality and timeliness are validated by rigorous and extensive experiments, providing concrete evidence of its superiority over conventional methods such as JPEG2000 \cite{christopoulos2000jpeg2000} with low-density parity check (LDPC) \cite{chen2005reduced}, CNN-based deep JSCC \cite{8723589}, and CDDM \cite{10480348}. The code is open-sourced at \url{https://github.com/JianhuaPei/LDM-enabled-SemCom-system}.

The rest of this paper is organized as follows. Section \ref{sec:System} briefly introduces the proposed wireless SemCom system model, existing challenges, and related works. Section \ref{sec:WGAN} elaborates on the JSCC design of the proposed SemCom system for transmitting data with unknown errors and distributions. The real-time channel denoising implementation is established by EECD in Section \ref{sec:LCDM}. Numerical experiments are given in Section \ref{sec:NE}. Section \ref{sec:Conclusion} concludes the paper. Supporting lemmas are included in the Appendix for reference.


\section{System Overview and Methodological Innovations}
\label{sec:System}

\subsection{Problem Formulation} 

Conventional DL-based JSCC typically consists of a semantic encoder $E_{\bm{\phi}}(\cdot)$ parameterized by $\bm{\phi}$ at the transmitter and a semantic decoder $G_{\bm{\psi}}(\cdot)$ parameterized by $\bm{\psi}$ at the receiver. The semantic encoder usually encodes the source data $\bm{x}$ into low-dimensional latent vectors $\bm{z}$ and transmits them over the wireless channel, and finally, the semantic decoder reconstructs the data based on the received signals. However, some DL-based JSCC systems face the following challenges:
\begin{enumerate}[leftmargin=12pt]
\item \textbf{Semantic Error:} Due to unreasonable photographing, storage, or cyber attacks, the transmitted data may contain some imperceptible errors or noises $\bm{\delta}$, which may cause DL-based communication systems to reconstruct data with semantic ambiguities based on the contaminated data $\bm{x}'=\bm{x}+\bm{\delta}$ at the receiver.
\item \textbf{Unknown Distribution:} When a DL-based communication system transmits data $\bm{x}''$ with unknown distribution, i.e., the data type is not included in the training dataset, the decoder may generate data with different semantics. 
\item \textbf{Channel Uncertainties:} The wireless channels are inevitably subject to varying fading gains and noises with uncertain SNRs. Assume that the transmitted complex latent signal is denoted by $\bm{z}_c \in \mathbb{C}^k$, and the latent vector needs to utilize the wireless channel by $k$ times to reach the receiver, where $k$ represents the size of latent space. At time $t$, the $i$-th symbol of complex $k$-length received noisy signal $\bm{z}'=\bm{y}_c$ can be represented as 
\begin{equation}
	\begin{aligned}
y_{c,i} = h_{c,i}z_{c,i}+n_{c,i}, 
\end{aligned}
\end{equation}
where $z_{c,i}$ represents the $i$-th component of $\bm{z}_c$, $h_{c,i} = \sum_{p=1}^P \alpha_p e^{-j2\pi f\tau_p (t)}$, $\alpha_p$ is the signal amplitude of the $p$-th path, $P$ denotes the number of paths, $f$ is the carrier frequency, $\tau_p(t)$ denotes the phase shift, $n_{c,i} \sim \mathcal{CN}(0, \sigma^2)$ represents the complex Gaussian noise. Considering the effects of multipath fading and scattering, $h_{c,i}$ are independent and identically distributed (i.i.d.) Rician fading gains, which is denoted by 
\begin{equation}
	\begin{aligned}
h_{c,i} = \sqrt{\frac{K}{K+1}}+\sqrt{\frac{1}{K+1}}h_{Rayleigh,i}, 
\end{aligned}
\end{equation}
where $h_{Rayleigh,i}$ are i.i.d. Rayleigh fading gains and $K$ is the ratio of direct radio waves' power and non-direct radio waves' power. When $K=\infty$, the wireless channel becomes additive white Gaussian noise (AWGN) channel, and the channel becomes Rayleigh channel when $K=0$. 
\end{enumerate}

\subsection{Related Works} 

Existing SemCom models primarily focus on the extraction and transmission of semantic information, with few methods addressing the vulnerability of DL-based SemCom systems to semantic errors \cite{adesina2022adversarial}. Current mainstream approaches in the fields of communication and AI for handling outliers in transmitted data still rely on anomaly detection \cite{liu2020deep} and data recovery \cite{10542391}. Therefore, there is a need for training strategies for a semantic encoder that is robust to semantic errors.

The handling of out-of-distribution data in DL-based SemCom systems has been a research hotspot. Common approaches include transfer learning \cite{9953099, 10416926}, ensemble learning \cite{nozza2016deep}, and multi-task training \cite{zhang2022unified}, all of which can enhance the semantic accuracy of decoded data following unknown distribution. However, these methods face bottlenecks such as high resource demands and long processing times, so strategies for quickly transmitting out-of-distribution data still need further exploration.

In \cite{yang2022semantic}, the main tasks of semantic communication systems include the extraction and transmission of semantic information. Some methods focusing on extracting semantic information do not account for channel uncertainties, while those that consider channel imperfections mainly focus on the JSCC approach \cite{8723589} or deploying denoisers at the receiver, such as denoising autoencoders \cite{khan2019robust}, conditional GANs \cite{ye2020deep}, and diffusion models \cite{zilberstein2024joint, 10480348}. However, these channel uncertainty mitigation mechanisms still face issues such as high latency and low precision.

\begin{figure*}[!t]
	\vspace{-0.2cm}
	\centerline{\includegraphics[width=0.98\textwidth]{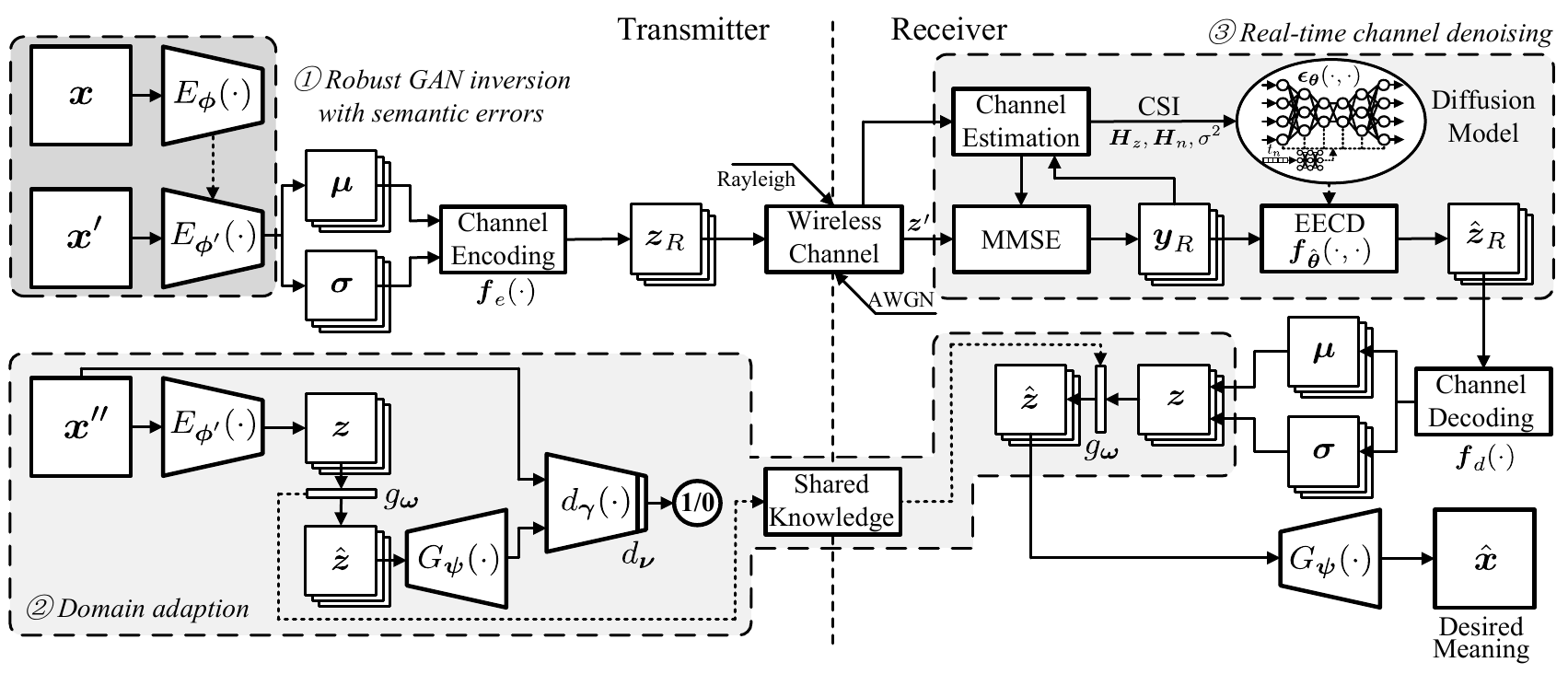}}
	\caption{The proposed SemCom system with three addressed DL-based communication challenges: \ding{172} robust GAN inversion with semantic errors, \ding{173} domain adaptation with unknown distribution, and \ding{174} real-time wireless channel denoising with EECD, where $\bm{\mu}$ and $\bm{\sigma}$ are the two components of latent bottleneck of VAE, $\bm{H}_z$, $\bm{H}_n$, $\sigma^2$, $\bm{z}_R$, and $\bm{y}_R$ are the CSIs, real-valued transmitted encodings, and equalized received signals, respectively, as defined in Section \ref{sec:LCDM}. $f_e(\cdot)$ represents the modulation encoding for 256-QAM, while $f_d(\cdot)$ represents the demodulation decoding for 256-QAM. Other symbols' definition can be found in Section \ref{sec:System}.}
	\label{JSCC_LDM}
	\vspace{-0.4cm}
\end{figure*}

\subsection{SemCom System Overview} 

The proposed LDM-enabled SemCom system is a JSCC approach that utilizes an additional diffusion model for channel denoising with quadrature amplitude modulation (QAM), as depicted in Fig. \ref{JSCC_LDM}. Specifically, JSCC consists of the encoder $E_{\bm{\phi}}(\cdot)$ with target distribution $q_{\bm{\phi}}(\bm{z}|\bm{x})$ at the transmitter, DM $\bm{\epsilon}_{\bm{\theta}}(\cdot,\cdot)$ parameterized by $\bm{\theta}$ with denoised latent vector's distribution $p_{\bm{\theta}}(\bm{z})$ at the receiver, and decoder $G_{\bm{\psi}}(\cdot)$ with reconstruction target distribution $q_{\bm{\psi}}(\bm{x}|\bm{z})$ by utilizing the synthesized encodings $\bm{z}$ from DM. The goal of training this LDM is to learn $\left \{ \bm{\phi}, \bm{\theta}, \bm{\psi} \right \}$ by minimizing the overall variational upper bound (VUB) \cite{vahdat2021score} to eliminate the gap between encodings of $E_{\bm{\phi}}(\cdot)$ and output of $\bm{\epsilon}_{\bm{\theta}}(\cdot,\cdot)$, and to ensure the quality of the decoded data $\hat{\bm{x}}$, defined as follows:
\begin{equation} \label{JSCC_loss}
	\begin{aligned}
&\mathcal{L}_{JSCC}\left ( \bm{\phi}, \bm{\theta}, \bm{\psi} \right ) \\
&= \mathbb{E}_{q_{\bm{\phi}}(\bm{z}|\bm{x})} \left [\mathcal{D}_{KL}\left ( q_{\bm{\phi}}(\bm{z}|\bm{x}) || p_{\bm{\theta}}(\bm{z}) \right )\right ] + \mathbb{E}_{q_{\bm{\phi}}(\bm{z}|\bm{x})}\left [ -\log q_{\bm{\psi}}(\bm{x}|\bm{z})\right ]   \\
& =   \underbrace{\mathbb{E}_{q_{\bm{\phi}}(\bm{z}|\bm{x})}\left [ \log q_{\bm{\phi}}(\bm{z}|\bm{x})\right ]}_{\textrm{transmitter encoding entropy}}  + \underbrace{\mathbb{E}_{q_{\bm{\phi}}(\bm{z}|\bm{x})} \left [  -\log p_{\bm{\theta}}(\bm{z})\right ]}_{\textrm{channel cross entropy}}\\
& \quad  +    \underbrace{\mathbb{E}_{q_{\bm{\phi}}(\bm{z}|\bm{x})}\left [ -\log q_{\bm{\psi}}(\bm{x}|\bm{z})\right ]}_{\textrm{receiver reconstruction term}},
\end{aligned}
\end{equation}
where $\mathcal{D}_{KL}(\cdot||\cdot)$ denotes the Kullback-Leibler divergence, and $q_{\bm{\phi}}(\bm{z}|\bm{x})$ approximates the true posterior $q_{\bm{\psi}}(\bm{z}|\bm{x})$ of decoder. The loss in Eq. \eqref{JSCC_loss} has been widely applied and validated in fast data generation \cite{rombach2022high}. Unlike data generation, the goal of SemCom systems is to make the reconstructed data in receivers show desired meaning. Consequently, Eq. \eqref{JSCC_loss} is divided into three terms: the encoding entropy term for semantic encoding at the transmitter, the cross entropy term for synthesized denoised bottlenecks $\bm{z}$ at the wireless channel, and the reconstruction term for perceptual quality at the receiver. As a result, define $\bm{x}'/\bm{x}''$ and $\bm{z}'$ as the transmitted data with aforementioned potential issues, the communication objective terms in Eq. \eqref{JSCC_loss} are rewritten as:
\begin{itemize}[leftmargin=10pt]
		\item[$\bullet$] Transmitter: $\mathbb{E}_{q_{\bm{\phi}}(\bm{z}|\bm{x}'/\bm{x}'')}\left [ \log q_{\bm{\phi}}(\bm{z}|\bm{x}'/\bm{x}'')\right ]$,
		\item[$\bullet$] Wireless channel: $\mathbb{E}_{q_{\bm{\phi}}(\bm{z}|\bm{x}'/\bm{x}'')} \left [  -\log p_{\bm{\theta}}(\bm{z}|\bm{z}')\right ]$,
		\item[$\bullet$] Receiver: $\mathbb{E}_{q_{\bm{\phi}}(\bm{z}|\bm{x}'/\bm{x}'')}\left [ -\log q_{\bm{\psi}}(\bm{x}/\bm{x}''|\bm{z})\right ]$.
\end{itemize}

The proposed SemCom system addresses the above three challenges of the DL-based communication system point-to-point. As detailed in Subsection \ref{dwgan}, the basic encoder-decoder architecture of the proposed system consists of a variational encoder $E_{\bm{\phi}}(\cdot)$ and generator $G_{\bm{\psi}}(\cdot)$ of WGAN. Based on this, as illustrated in Fig. \ref{JSCC_LDM}, the threefold improvements are further clarified as follows:
\begin{enumerate}[leftmargin=12pt]
	\item \textbf{Robust GAN Inversion:} The imperceptible semantic error that leads to the maximum reconstruction error in the DL-based SemCom system is defined and obtained through adversarial convex optimization. Based on this semantic error, the parameters of the optimized robust encoder are updated from $\bm{\phi}$ to $\bm{\phi}'$ to encode normal latent space for transmission. The specific GAN robust inversion method is detailed in Subsection \ref{robustencoder}.
	\item \textbf{Domain Adaptation:} When transmitting out-of-distribution data, the lightweight single-layer $g_{\bm{\omega}}$ and $d_{\bm{\nu}}$ are exploited for one-shot fast and adversarial domain adaptation learning. The learned parameters $\bm{\omega}$ will be seamlessly transmitted to the receiver along with the data for latent space transformation, and the decoder will ultimately output semantically consistent data. The specific implementation can be found in Subsection \ref{ofdls}.
	\item \textbf{Low-Latency Channel Denoising:} Assuming that the CSIs are known, EECD is proposed to distill LDM from a multi-step denoising process into one setp, thereby reducing the computational complexity of online sampling during real-time communication. The detailed wireless channel modeling, training and sampling approaches of the latent channel denoising DM, and one-step real-time channel denoising algorithm are elucidated in Section \ref{sec:LCDM}.
\end{enumerate}
These advancements open up a range of potential applications, including real-time video streaming, remote healthcare monitoring, and intelligent transportation systems, where low-latency and high-quality communication is crucial. Moreover, the ability to effectively manage semantic ambiguities and wireless channel noise further positions this system as a valuable solution in IoT environments and augmented reality applications, where accurate and timely semantic information exchange is essential.

\section{Deep JSCC for data with unknown errors and distributions}
\label{sec:WGAN}
In this section, the proposed robust and high-quality JSCC is further detailed. In Subsection \ref{dwgan}, WGAN and its inversion network are introduced to serve as the decoder and encoder. Subsection \ref{robustencoder} then provides an fine-tuning encoder that is robust to errors. The fast and reliable SemCom approach for data of unknown distribution is implemented in Subsection \ref{ofdls} by latent space exploration for unknown distributions.

\subsection{Decoder and Encoder: GAN and GAN Inversion} \label{dwgan}
The generators of GANs with slightly lower generation quality than DMs are still selected as the semantic decoder for the proposed LDM-enabled JSCC for its single-step data generation property. GAN is formulated based on zero-sum game between a discriminator $D_{\bm{\gamma}}(\cdot)$ and a generator $G_{\bm{\psi}}(\cdot)$ with the adversarial training objective given as follows:
\begin{equation}
	\begin{aligned}
\min_{\bm{\psi}}\max_{\bm{\gamma}} & \mathbb{E}_{q(\bm{x})}\left [ \log D_{\bm{\gamma}}(\bm{x}) \right ] + \mathbb{E}_{ q_{\bm{\psi}(\bm{z})}}\left [ \log \left ( 1- D_{\bm{\gamma}}(G_{\bm{\psi}}(\bm{z}))\right )  \right ], 
\end{aligned}
\end{equation}
where $q(\bm{x})$ denotes the distribution of input data, $q_{\bm{\psi}}(\bm{z})$ is the prior distribution of latent vector $\bm{z}$ and $q_{\bm{\psi}}(\bm{z}) = \mathcal{N}(\bm{0}, \bm{I})$ in GANs. Furthermore, to overcome these challenges of training instability and collapse mode, WGAN \cite{adler2018banach} is established by replacing $\mathcal{D}_{KL}$ and Jensen-Shannon divergence $\mathcal{D}_{JS}$ with Wassertein distance $\mathcal{D}_{\mathcal{W}}$. Similarly, other variants of GAN that have been proposed for better perceptual reconstruction can also be utilized as the JSCC decoder. Among them, StyleGAN \cite{abdal2019image2stylegan} and Diff-GAN distillated from DMs \cite{wang2022diffusion} have achieved impressive generation results, and Diff-GAN is even comparable to the DMs in some datasets. 

Compared to VAE, GANs are skilled in generating data with high-resolution. Nonetheless, the task of SemCom is to ensure that the signals received by receivers can accurately convey the meaning, while minimizing the bandwidth of SemCom. For this reason, JSCC requires an encoder to determine the latent bottlenecks of the transmitted data, also known as GAN inversion \cite{xia2022gan}. Commonly, the solution of GAN inversion is to utilize a neural network-based encoder to find the optimal latent vector $\bm{z}$ given the transmitted data $\bm{x}$. 
\begin{proposition}
Ignoring the channel's cross entropy term of the latent space and taking into account the receiver reconstruction term and the transmitter encoding entropy, the VUB defined in Eq. \eqref{JSCC_loss} can be transformed into 
\begin{equation} \label{LJSCC2}
	\begin{aligned}
& \mathcal{L}'_{JSCC} =\mathbb{E}_{ q_{\bm{\phi}}(\bm{z}|\bm{x})}\left [ -\log p_{\bm{\psi}}(\bm{x}|\bm{z}) \right ]+ \mathbb{E}_{ q_{\bm{\phi}}(\bm{z}|\bm{x})}\left [ \log q_{\bm{\phi}}(\bm{z}|\bm{x}) \right ]\\
& \geq \mathbb{E}_{ q_{\bm{\psi}}(\bm{z})}\left [ -\log p_{\bm{\psi}}(\bm{x}) \right ] + \mathbb{E}_{ q(\bm{x})} \left [  \mathcal{D}_{KL}(  q_{\bm{\phi}}(\bm{z}|\bm{x}) \parallel p_{\bm{\psi}}(\bm{z}|\bm{x}) )  \right ]\\
& \geq \mathbb{E}_{ q_{\bm{\psi}}(\bm{z})}\left [ -\log p_{\bm{\psi}}(\bm{x}) \right ],
	\end{aligned}
\end{equation}
where the proof can be seen in Appendix \ref{transVAEWGAN}.
\end{proposition}
Apparently, the term $\mathbb{E}_{ q_{\bm{\psi}}(\bm{z})}\left [ -\log p_{\bm{\psi}}(\bm{x}) \right ]$ in Eq. \eqref{LJSCC2} can be replaced with the training objective of the generator $G_{\bm{\psi}}(\cdot)$ of WGAN, and term $\mathbb{E}_{ q(\bm{x})} \left [  \mathcal{D}_{KL}(  q_{\bm{\phi}}(\bm{z}|\bm{x}) \parallel p_{\bm{\psi}}(\bm{z}|\bm{x}) )  \right ]$ indicates that the encoding latent vector $\bm{z}$ should be as consistent as possible with the input latent space of generator $G_{\bm{\psi}}(\cdot)$ under the same transmitted data $\bm{x}$, which can be addressed by training VAE. When DM can generate realistic $\bm{z}$ as much as possible, jointly or separately training VAE and WGAN is equivalent to minimizing the loss $\mathcal{L}_{JSCC}$. The output of VAE’s encoder can be represented as $q_{\bm{\phi}}(\bm{z}|\bm{x})\sim \mathcal{N}(\bm{\mu}, \bm{\sigma}^2 )$, and $\bm{z}$ is reparameterized as $\bm{z} = \bm{\mu} + \bm{\sigma} \odot \bm{\epsilon}  $, where $\bm{\epsilon}  \sim \mathcal{N}(\bm{0},\bm{I})$ and $\odot$ denotes the element-wise product. Consequently, by combining the decoupled optimization objectives of WGAN and Eq. \eqref{LJSCC2}, the training process of deep CNN based VAE-WGAN can be found in \cite{larsen2016autoencoding} and Appendix \ref{trainingVAEWGANGP}.

\begin{figure}[!h]
	\vspace{-0.1cm}
	\centerline{\includegraphics[width=0.5\textwidth]{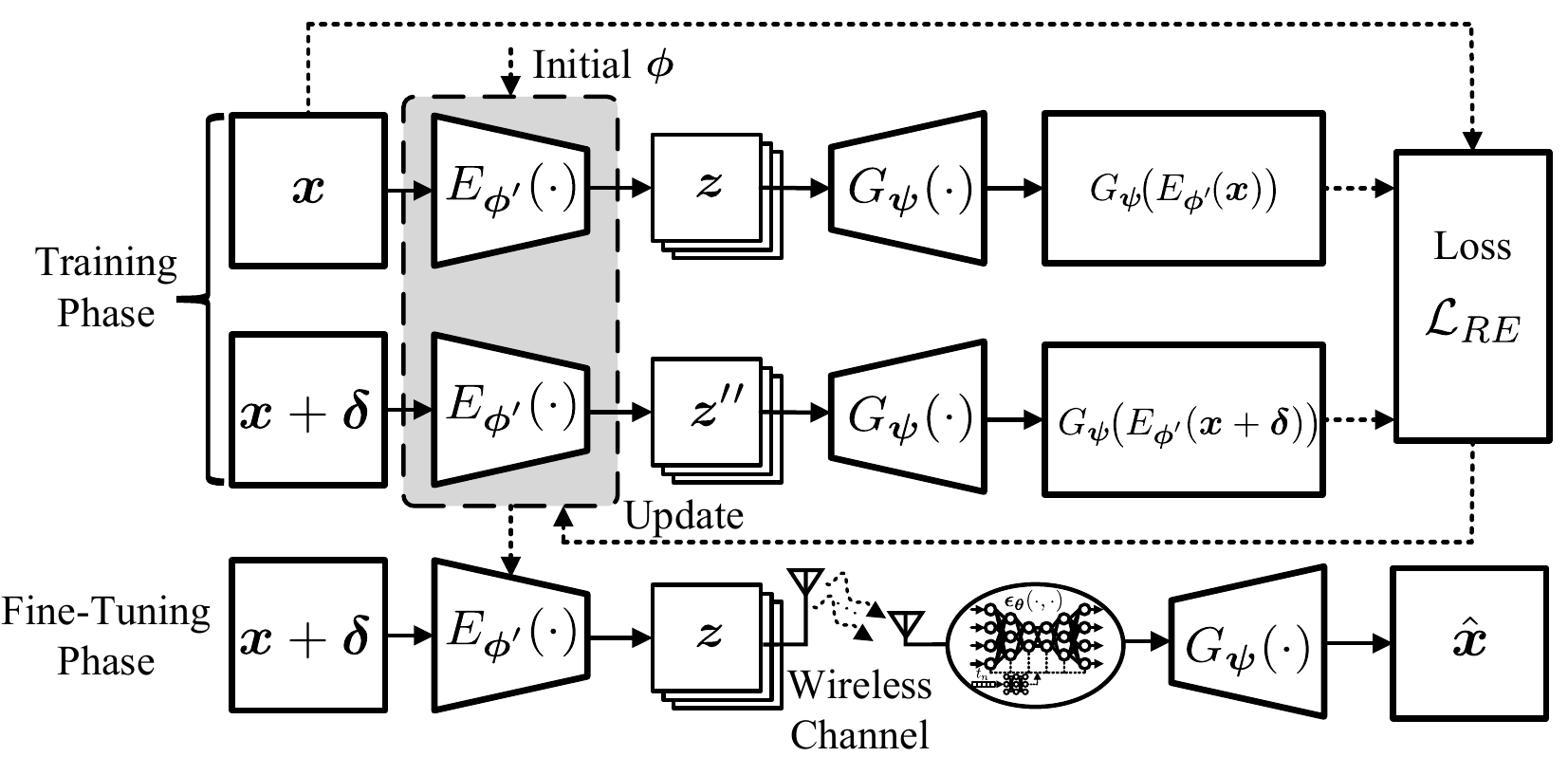}}
	\caption{Self-supervised robust encoder optimization with semantic error $\bm{\delta}$.}
	\label{Task1}
	\vspace{-0.0cm}
\end{figure}

\subsection{Robust Semantic Encoder} \label{robustencoder}
As discussed in Section \ref{sec:System}, VAE-WGAN based SemCom systems suffer from inevitable vulnerabilities. The adversarial attack methods explore the vulnerabilities of neural networks utilized for classification and regression tasks, with the goal of determining a sufficiently small and unnoticed error $\bm{\delta}$ to mislead the classification or regression results. The unified optimization objective of adversarial attacks is given by
\begin{equation} \label{adverattack}
	\begin{aligned}
\min_{\bm{\delta}} \quad &\bm{d}(\bm{x}, \bm{x}+\bm{\delta})\\
\textrm{s.t.:}  \quad &\bm{f}(\bm{x}+\bm{\delta}) = \mathcal{T}, \quad \bm{L} \leq \bm{x}+\bm{\delta} \leq \bm{U},
\end{aligned}
\end{equation}
where $\bm{d}(\cdot,\cdot)$ is the distance function that measure the differences between two data points, $\bm{f}(\cdot)$ is the attacked nerual network, $\mathcal{T}$ denotes the output of the DL model, $\bm{L}$ and $\bm{U}$ represent the physical lower and upper bounds of input data $\bm{x}$, respectively. Specifically, in classification tasks, $\mathcal{T}$ is the class that is inconsistent with original category of $\bm{x}$, e.g., the hackers can exploit objective \eqref{adverattack} to make their cyber attacks undetectable by network $\bm{f}(\cdot)$. In regression tasks, $\mathcal{T}$ represents the output data that is not the same as the original regression results, e.g., when transmitting a digital image in SemCom systems, decoder may reconstruct another type of digital image with semantic ambiguities.

\begin{algorithm}[!t]  	\label{trainingrobustencoder}
	\small
	\caption{Training algorithm of robust GAN inversion $E_{\bm{\phi}'}(\cdot)$}
		\LinesNumbered
		\KwIn{ Dataset $q(\bm{x})$, learning rate $\eta_1$ and $\eta_2$, original encoder $E_{\bm{\phi}}(\cdot)$, generator $G_{\bm{\psi}}(\cdot)$, }
		\KwOut{The updated robust encoder $E_{\bm{\phi}'}(\cdot)$}
		\textbf{Initialize} $\bm{\phi}' \leftarrow \bm{\phi}$\;
		\Repeat{Converged}{
			Sample $\bm{x} \sim q(\bm{x})$\;
			Initialize $\bm{\delta}^0 \leftarrow \bm{0}$ and $i \leftarrow 1$\;
			\Repeat{Converged}{
            Compute $\bm{\delta}^i  \leftarrow P_C\left (\bm{\delta}^{i-1}- \eta_1 \nabla_{\bm{\delta}} \bm{e}(\bm{\delta}^{i-1})\right ) $\;
			Update $i \leftarrow i+1$\;
			}
		Determine $\bm{\delta}$ by $\bm{\delta} \leftarrow \bm{\delta}^k$\;
		Update $\bm{\phi}'$ by $\bm{\phi}' \leftarrow \bm{\phi}'-\eta_2 \nabla_{\bm{\phi}'}\Big [ \mathbb{E}_q \big ( \bm{d}(\bm{x}, G_{\bm{\psi}}(E_{\bm{\phi}'}(\bm{x}))) + \bm{d}(G_{\bm{\psi}}(E_{\bm{\phi}'}(\bm{x})), G_{\bm{\psi}}(E_{\bm{\phi}'}(\bm{x}+\bm{\delta}))) \big ) \Big ]$\;
		}
		\textbf{Return} Robust GAN inversion $E_{\bm{\phi}'}(\cdot)$
\end{algorithm}

In order to address the challenges of semantic errors, SemCom systems should have a robust and enhanced encoder that can handle those outliers. The objective for the sufficiently small semantic error $\bm{\delta}$ constrained by $\varepsilon$ that leads to the maximum receiver reconstruction error is given by 
\begin{equation} \label{maxdelta2}
	\begin{aligned}
\max_{\bm{\delta}} \quad &\bm{d} \left (G_{\bm{\psi}}(E_{\bm{\phi}}(\bm{x})), G_{\bm{\psi}}(E_{\bm{\phi}}(\bm{x}+\bm{\delta})) \right )\\
\textrm{s.t.:}  \quad &E_{\bm{\phi}}(\bm{x}+\bm{\delta}) \sim \mathcal{N}(\bm{0},\bm{I}), \left \|  \bm{\delta} \right \|_p \leq \varepsilon , \bm{L} \leq \bm{x}+\bm{\delta} \leq \bm{U},
\end{aligned}
\end{equation}
where $\left \| \cdot \right \|_p$ denotes the p-norm. In this way, objective \eqref{maxdelta2} simultaneously utilizes the vulnerabilities of both the encoder $E_{\bm{\phi}}(\cdot)$ and generator $G_{\bm{\psi}}(\cdot)$. When solving \eqref{maxdelta2}, its objective can be transformed into a standard convex optimization problem 
\begin{equation} 
	\begin{aligned}
\min_{\bm{\delta}} \quad & \underbrace{\lambda\left \|  \bm{\delta} \right \|_p - \bm{d} \left (G_{\bm{\psi}}(E_{\bm{\phi}}(\bm{x})), G_{\bm{\psi}}(E_{\bm{\phi}}(\bm{x}+\bm{\delta})) \right )}_{\bm{e}(\bm{\delta})}\\
\textrm{s.t.:}  \quad &E_{\bm{\phi}}(\bm{x}+\bm{\delta}) \sim \mathcal{N}(\bm{0},\bm{I}), \bm{L} \leq \bm{x}+\bm{\delta} \leq \bm{U},
\end{aligned}
\end{equation}
where $\lambda$ is the penalty coefficient. The constrained convex optimization problem can be solved by using the projected gradient descent (PGD) \cite{lanfredi2023quantifying} iterative optimization method to obtain semantic error $\bm{\delta}$. Consequently, the semantic error at the i-th iteration $\bm{\delta}^i$ is denoted by
\begin{equation} 
	\begin{aligned}
\bm{\delta}^i = P_C\left (\bm{\delta}^{i-1}- \eta \nabla_{\bm{\delta}} \bm{e}(\bm{\delta}^{i-1})\right )= P_C(\bm{\varsigma}^{i}),
	\end{aligned}
\end{equation}		
where $P_C(\bm{\varsigma}^{i})$ represents the projection of $\bm{e}(\bm{\delta})$ on the set of constraints $C$, i.e., $\bm{\delta}^i = P_C(\bm{\varsigma}^i):= {\arg \min}_{\bm{\delta}\in C}\frac{1}{2}\left \| \bm{\delta} -\bm{\varsigma}^i \right \|_2^2$. 

\begin{proposition}
Let $\bm{z}''$ be the erroneous latent vector encoded from data containing semantic errors $\bm{x}'=\bm{x}+\bm{\delta}$, the robust VUB for semantic errors is defined as 
\begin{equation} \label{vlbr}
	\begin{aligned}
& \mathbb{E}_q\left [ - \log p_{\bm{\psi}}(\bm{x}) \right ] = \mathbb{E}_q\left [  -\log \int p_{\bm{\psi}}(\bm{x}, \bm{x}+\bm{\delta}) d(\bm{x}+\bm{\delta}) \right ]\\
&  \leq \mathbb{E}_{q(\bm{z})}\left [-\log p_{\bm{\psi}}(\bm{x}|\bm{z})\right ] + \mathbb{E}_{q(\bm{z})}\left [-\log p_{\bm{\psi}}(\bm{z})\right ]\\
& +\mathbb{E}_{q(\bm{z}'')}\left [-\log p_{\bm{\psi}}(\bm{z}'')\right ]+ \frac{\beta}{2}\mathbb{E}_{q(\bm{z}, \bm{z}'')}\bm{d}(\bm{z}, \bm{z}'')\\
&  -\mathbb{E}_{q(\bm{z}, \bm{z}'')} \left [ -\log q(\bm{z}, \bm{z}'')\right ],
	\end{aligned}
\end{equation}
where $q(\bm{z}, \bm{z}'')$ denotes the joint distribution and the proof can be seen in Appendix \ref{proofVUB}.
\end{proposition}
Evidently, the first and second terms of Eq. \eqref{vlbr} have been addressed in VAE-WGAN based JSCC, and the third term has also been optimized by sloving the semantic errors $\bm{\delta}$. For this reason, the training objective of robust encoder is
\begin{equation} \label{simplevlbr}
	\begin{aligned}
\min_{\bm{\phi}'} \quad \frac{\beta}{2} \mathbb{E}_{q(\bm{z}, \bm{z}'')}\bm{d}(\bm{z}, \bm{z}'')+ \mathbb{E}_{q(\bm{z}, \bm{z}'')} \left [ \log q(\bm{z}, \bm{z}'')\right ],
\end{aligned}
\end{equation}
where $\bm{\phi}'$ denotes the robust encoder parameters. In \cite{cemgil2019adversarially}, the objective \eqref{simplevlbr} is equivalent to minimizing the Wassertein distance between $\bm{z}$ and the incorrect $\bm{z}''$. Nevertheless, the ultimate goal of SemCom is to accurately reconstruct the transmitted data. Therefore, as the parameter $\bm{\psi}$ is fixed, the optimal robust encoder parameters $\bm{\phi}'$ considering both encoder and decoder vulnerabilities is 
\begin{equation} 
	\begin{aligned}
\bm{\phi}'= & \mathop{\arg \min}_{\bm{\phi}'} \mathcal{L}_{RE} =  \mathop{\arg \min}_{\bm{\phi}'}\mathbb{E}_q \Big [ \bm{d}(\bm{x}, G_{\bm{\psi}}(E_{\bm{\phi}'}(\bm{x})))  \\
&+ \bm{d}(G_{\bm{\psi}}(E_{\bm{\phi}'}(\bm{x})), G_{\bm{\psi}}(E_{\bm{\phi}'}(\bm{x}+\bm{\delta})))\Big ].
\end{aligned}
\end{equation}
In summary, the self-supervised training process of robust encoder with prior VAE-WGAN is depicted in Fig. \ref{Task1} and illustrated in Algorithm \ref{trainingrobustencoder}.

\subsection{Out-of-Domain Latent Space} \label{ofdls}
DL-based SemCom will significantly degrade their performances when facing data types that are not included in the training dataset (\textit{out-of-domain}). For the proposed JSCC approach, when the transmitter wants to send an \textit{out-of-domain} data, the robust encoder $E_{\bm{\phi}'}(\cdot)$ may encode an abnormal latent vector, and the decoder at the receiver will reconstruct data that is semantically different from the transmitted data. For this reason, when facing the data with unknown distributions, SemCom system should improve its generalization abilities and be able to quickly and dynamically adapt to search for the optimal \textit{out-of-domain} latent space. 

\begin{figure}[!h]
	\vspace{-0.3cm}
	\centerline{\includegraphics[width=0.5\textwidth]{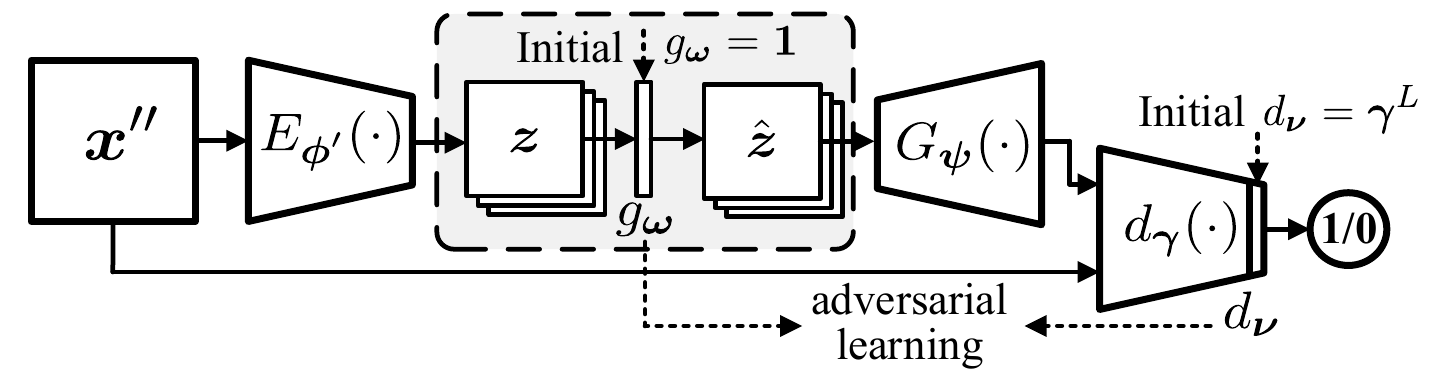}}
	\caption{\textit{Out-of-domain} latent space determination using lightweight single-layer network and adversarial training method.}
	\label{Task2}
	\vspace{-0.0cm}
\end{figure}

To address this issue, a learning-based adaptor constructed by a lightweight single-layer neural network is utilized for \textit{out-of-domain} latent space determination. Considering the characteristics of VAE-WGAN, as shown in Fig. \ref{Task2}, the adapter $g_{\bm{\omega}}(\cdot)$ parameterized by $\bm{\omega}$ is placed between the robust encoder and generator. Subsequently, when the transmitted data follows an unknown distribution, adaptor $g_{\bm{\omega}}(\cdot)$ can perform one-shot learning to transform the latent vector $\bm{z}$ encoded by $E_{\bm{\phi}'}(\cdot)$ into
\begin{equation} 
	\begin{aligned}
\hat{\bm{z}} = g_{\bm{\omega}}(\bm{z}) = \bm{\omega}^{\top}\bm{z}+ \bm{b},
\end{aligned}
\end{equation}
where $\bm{b}$ denotes the bias of adaptor $g_{\bm{\omega}}(\cdot)$. In order to improve the data quality of reconstruction, inspired by the adversarial training startegy of WGAN, this paper considers another adaptor $d_{\bm{\nu}}(\cdot)$ composed of a lightweight fully connected (FC) layer for adversarial training with $g_{\bm{\omega}}(\cdot)$. As illustrated in Fig. \ref{Task2}, during online training, the FC layer of discriminator $D_{\bm{\gamma}}(\cdot)$ is replaced by $d_{\bm{\nu}}(\cdot)$, and the original discriminator after removing the FC layer is denoted by $d_{\bm{\gamma}}(\cdot)$. In this way, the online training process of $g_{\bm{\omega}}(\cdot)$ is similar to WGAN's training approach as illustrated in Algorithm \ref{trainingadaptor}.

\begin{algorithm}[!t]  	\label{trainingadaptor}
	\small
	\caption{Online training algorithm of \textit{out-of-domain} adaptor $g_{\bm{\omega}}(\cdot)$}
		\LinesNumbered
		\KwIn{ Data following unknown distribution $q(\bm{x}'')$, learning rate $\eta$,  gradient penalty coefficient $\lambda$, robust encoder $E_{\bm{\phi}'}(\cdot)$, generator $G_{\bm{\psi}}(.)$, discriminator $D_{\bm{\gamma}}(\cdot)$ and the parameters of last layer (L-th layer) is denoted by $\bm{\gamma}^L$}
		\KwOut{The online-updated adaptor $g_{\bm{\omega}}(\cdot)$}
		\textbf{Initialize} $\bm{\omega} \leftarrow \bm{1}$ and $\bm{\nu} \leftarrow \bm{\gamma}^L$\;
		\Repeat{Converged}{
			Sample $\bm{x}'' \sim q(\bm{x}'')$, $\bm{z} \sim q_{\bm{\psi}}(\bm{z})$, and $\epsilon \sim U[0,1]$\;
                Compute $\hat{\bm{x}} \leftarrow \epsilon\bm{x}'' + (1-\epsilon)G_{\bm{\psi}}(g_{\bm{\omega}}(\bm{z}))$\;
			Update $\bm{\nu} \leftarrow \bm{\nu} -\eta \nabla_{\bm{\nu}}\Big [ \mathbb{E}_q \big (- d_{\bm{\nu}}(d_{\bm{\gamma}}(\bm{x}'')) +  d_{\bm{\nu}}(d_{\bm{\gamma}}(G_{\bm{\psi}}(g_{\bm{\omega}}(\bm{z})))) + \lambda (\left \| \nabla_{\hat{\bm{x}}} d_{\bm{\nu}}(d_{\bm{\gamma}}(\hat{\bm{x}}))  \right \|_2 -1)^2 \big ) \Big ]$\;
			Update $\bm{\omega} \leftarrow \bm{\omega}-\eta \nabla_{\bm{\omega}}\Big [  \mathbb{E}_q \left ( -d_{\bm{\nu}}(d_{\bm{\gamma}}(G_{\bm{\psi}}(g_{\bm{\omega}}(\bm{z})))) \right )\Big ]$\;
		}
		\textbf{Return} the parameters $\bm{\omega}$ of adaptor $g_{\bm{\omega}}(\cdot)$
\end{algorithm}

In summary, when the SemCom system transmits data \textit{in domain}, the parameters $\bm{\omega}$ of $g_{\bm{\omega}}(\cdot)$ is equal to $\bm{1}$, while transmitting data with significant reconstruction errors inferred by the decoder deployed at the transmitter, online learning Algorithm \ref{trainingadaptor} will be activated to improve the decoded data's quality. Due to the limited amount of training data and the fact that the given initial values of $\hat{\bm{z}}$ and $\bm{\omega}$ are close to the optimal values, this online-updated process will be very fast. When the online-learning is completed, in order to not change the weights $\bm{\phi}'$, $\bm{\psi}$, and $\bm{\theta}$ of robust encoder, generator, and LDM utilized for channel denoising, the adaptor is deployed at the receiver and defined as a dynamic lightweight nerual network. In other words, the parameters of the implemented adaptor $g_{\bm{\omega}}(\cdot)$ can be dynamically changed in the proposed SemCom system as shown in Fig. \ref{JSCC_LDM}. Ultimately, the semantically consistent \textit{out-of-domain} data is reconstructed according to $\hat{\bm{z}} = g_{\bm{\omega}}(\bm{z})$.

\section{Latent Channel Denoising Diffusion Model}
\label{sec:LCDM}
In this section, the wireless channel equalization under different conditions is firstly established in Subsection \ref{WCE}. The training objectives of original LDMs based on the received signals and its one-step real-time implementation are introduced in Subsections \ref{LDM} and \ref{EECD}.

\subsection{Wireless Channel Equalization} \label{WCE}
Optionally, minimum mean square error (MMSE) \cite{10185192} is usually utilized as a method for received signals equalization to avoid errors and improve efficiency. Consequently, let $\bm{h}_c = [h_{c,1}, \cdots, h_{c,k}]$ and $\bm{n}_c = [n_{c,i}, \cdots, n_{c,k}]$ be the channel state and noises, the addressed signals for the received signals $\bm{z}'=\bm{y}_c$ defined in Section \ref{sec:System} can be denoted by 
\begin{equation}
	\begin{aligned}
\bm{y}_{eq} & = \left (\bm{h}_c^H\bm{h}_c + \sigma^2\bm{I} \right )^{-1}\bm{h}_c^H \left (\bm{h}_c \bm{z}_c+ \bm{n}_c \right ) \\
& =	\left (\bm{h}_c^H\bm{h}_c + \sigma^2\bm{I} \right )^{-1}\bm{h}_c^H\bm{h}_c\bm{z}_c + \left (\bm{h}_c^H\bm{h}_c + \sigma^2\bm{I} \right )^{-1}\bm{h}_c^H \bm{n}_c.
	\end{aligned}
\end{equation}

For simplicity, the transmitted complex signals $\bm{z}_c$ can also be rewritten as $\bm{z}_R \in \mathbb{R}^{2k}$ in real-valued symbols, the output of equalization can be also decoupled as corresponding real-valued $\bm{y}_R \in \mathbb{R}^{2k}$. In this way, the 1-st to $k$-th components of $\bm{y}_R$ are 
\begin{equation}
	\begin{aligned}
y_{R,i} = \frac{\left | h_{c,i} \right |^2}{\left | h_{c,i} \right |^2 + \sigma^2} z_{R,i} + \frac{Re(h^H_{c,i})}{\left | h_{c,i} \right |^2 + \sigma^2}\sigma \epsilon, 
	\end{aligned}
\end{equation}
where $\epsilon \sim \mathcal{N}(0,1)$. And the $k+1$-th to $2k$-th components can defined as 
\begin{equation}
	\begin{aligned}
y_{R,i} = \frac{\left | h_{c,i} \right |^2}{\left | h_{c,i} \right |^2 + \sigma^2} z_{R,i} + \frac{Im(h^H_{c,i})}{\left | h_{c,i} \right |^2 + \sigma^2}\sigma \epsilon. 
	\end{aligned}
\end{equation}
To this end, the known diagonal CSI matrix $\bm{H}_z$ and noise coefficient matrix $\bm{H}_n$ can be defined as 
\begin{equation}
	\begin{aligned}
\bm{H}_z = diag \bigg ( &\frac{\left | h_{c,1} \right |^2}{\left | h_{c,1} \right |^2 + \sigma^2}, \cdots, \frac{\left | h_{c,k} \right |^2}{\left | h_{c,k} \right |^2 + \sigma^2},\\
 &\frac{\left | h_{c,1} \right |^2}{\left | h_{c,1} \right |^2 + \sigma^2}, \cdots, \frac{\left | h_{c,k} \right |^2}{\left | h_{c,k} \right |^2 + \sigma^2}     \bigg ),
\end{aligned}
\end{equation}
\begin{equation}
	\begin{aligned}
\bm{H}_n = diag  \bigg (  & \frac{Re(h_{c,1}^H) }{ \left | h_{c,1} \right |^2 + \sigma^2}, \cdots, \frac{Re(h_{c,k}^H)}{\left | h_{c,k} \right |^2 + \sigma^2}, \\
& \frac{Im(h_{c,1}^H)}{\left | h_{c,1} \right |^2 + \sigma^2}, \cdots, \frac{Im(h_{c,k}^H)}{\left | h_{c,k} \right |^2 + \sigma^2}  \bigg  ).
	\end{aligned}
\end{equation}
As a consequence, the conditional distribution of $\bm{y}_R$ under the estimated wireless CSI, i.e., $\bm{h}_c$ and SNRs, is 
\begin{equation} \label{conditionald}
	\begin{aligned}
q_{\textrm{MMSE}}\left (  \bm{y}_R | \bm{z}_R, \bm{H}_z, \bm{H}_n\right )= \mathcal{N}\left ( \bm{y}_R; \bm{H}_z\bm{z}_R, \bm{H}^2_{n}\sigma^2\bm{I}  \right ),
\end{aligned}
\end{equation}
which means the received signals is affected by the channel's fading gains and noises. Especially, $\bm{H}_z = \bm{H}_n =\bm{I}\in \mathbb{R}^{2k\times 2k}$ under AWGN channel.

\subsection{Multi-Step Latent Diffusion Model} \label{LDM}
The denoising task of receiver is to find the original transmitted signals $\bm{z}_R$ from transmitter given $\bm{y}_R$ and CSI. Accordingly, let $\bm{z}_0 = \bm{H}_z\bm{z}_R$, the cross-entropy term in SemCom system model can be transformed from $\mathbb{E}_{q}\left [  - \log p_{\bm{\theta}}( \bm{z}| \bm{z}')\right ]$ to $\mathbb{E}_q \left [ - \log p_{\bm{\theta}}(\bm{z}_0 | \bm{y}_R, \bm{H}_z, \bm{H}_n  ) \right ]$. LDM is selected for wireless channel denoising as it has powerful capabilities to generate realistic data and much lower computational complexity than the original DMs. Let $\left \{ \bm{z}_t \right \}^{t=T}_{t=0}$ be the noisy latent bottlenecks containing noises of different SNRs in the continuous time domain $t \in [0,T]$, where $\bm{z}_0$ is the starting latent vector. LDM defines a forward process through a unified stochastic differential equation (SDE)
\begin{equation}
	\begin{aligned}
d\bm{z} = \bm{u}(\bm{z},t)dt + \bm{g}(t)d\bm{w}_t,
	\end{aligned}
\end{equation}
where $\bm{u}(\bm{z},t)$ and $\bm{g}(t)$ are the drift and diffusion coefficients, and $\bm{w}_t$ is a standard Brownian motion. By considering the reverse process of SDE, the marginal distribution $p(\bm{z}_t)$ follows the solution trajectory of the probability flow-ordinary differential equation (PF-ODE)
\begin{equation} \label{PF-ODE}
	\begin{aligned}
d\bm{z} = \left [\bm{u}(\bm{z},t) - \frac{1}{2} \bm{g}^2(t) \nabla_{\bm{z}}\log p(\bm{z}_t)\right ]dt,
	\end{aligned}
\end{equation}
where $\nabla_{\bm{z}}\log p(\bm{z}_t)$ denotes the score function. Accordingly, similar to Elucidated DM (EDM) \cite{karras2022elucidating}, considering the conditional distribution in Eq. \eqref{conditionald}, this paper sets $\bm{u}(\bm{z},t)=0$, $\bm{g}(t)= \sqrt{2t\bm{H}_n}$, and $\bm{\sigma}(t)= \bm{H}_nt$, where $t \in [0,T]$. When solving the reverse sampling trajectory, $t$ requires a discrete schedule $\left \{ t_n\right \}^{n=N}_{n=0}$. Concretely, when $n=0$, $t_0 = 0$, and when $n \geq 1$, $t_n = \left ( t_1^{1/\rho} + \frac{n-1}{N-1} \left ( t_N^{1/\rho} - t_1^{1/\rho} \right )\right )^{\rho}$, where $\rho>0$. Moreover, unlike denoising diffusion probabilistic model (DDPM) \cite{ho2020denoising}, the utilized diffusion model adopts variance explosion (VE) strategy, and its associated forward process $\left \{ \bm{z}_t \right \}^{t=T}_{t=0}$ can be written as  
\begin{equation}
	\begin{aligned}
q\left ( \bm{z}_t|\bm{z}_0 \right ) = \mathcal{N}\left (\bm{z}_t; \bm{z}_0, t^2\bm{H}^2_n \bm{I} \right ).
	\end{aligned}
\end{equation}

In the reverse process, the denoising U-Net is usually utilized to predict approximation function $\bm{s}_{\bm{\theta}}(\bm{z},t)$ to approximate the score function $\nabla_{\bm{z}}\log p(\bm{z}_t)$. Noise prediction model $\bm{\epsilon}_{\bm{\theta}}(\bm{z}_t,t)$ is one of the most popular implementations of diffusion models, and $\bm{s}_{\bm{\theta}}(\bm{z},t) = -\frac{\bm{\epsilon}_{\bm{\theta}}(\bm{z}_t,t)}{\bm{H}_nt}$ \cite{zhou2023fast}. As a consequence, the tranining objective of LDM is to minimize the distance between noise prediction $\bm{\epsilon}_{\bm{\theta}}(\bm{z}_t,t)$ and actual noise $\bm{\epsilon}$ \cite{zhou2023fast}
\begin{equation}
	\begin{aligned}
& \mathcal{L}_{LDM}  = \mathbb{E}_q \left [ \left \|\bm{s}_{\bm{\theta}}(\bm{z},t) - \nabla_{\bm{z}}\log p(\bm{z}_t)  \right \|^2_2 \right ] \\
& = \mathbb{E}_{\bm{z}_R, \bm{\epsilon}_1, n} \left [ \left \|    
 \frac{\bm{\epsilon}_{\bm{\theta}} (\bm{H}_z\bm{z}_R+\bm{H}_nt_n\bm{\epsilon}_1, t_n)}{\bm{H}_n t_n} - \frac{\bm{\epsilon}}{\bm{H}_n t_n}  \right \|^2_2 \right ]\\
& \Leftrightarrow \mathbb{E}_q \left [ \left \| \bm{\epsilon}_{\bm{\theta}} (\bm{z}_t, t) - \bm{\epsilon}\right \|^2_2 \right ],
	\end{aligned}
\end{equation}
where $\bm{\epsilon}_1 \sim \mathcal{N}(\bm{0},\bm{I})$ and $ n\sim \mathcal{U}[1,N]$. Considering aforementioned conditions and settings, the PF-ODE defined in Eq. \eqref{PF-ODE} can be rewritten as 
\begin{equation} \label{PF-ODE2}
	\begin{aligned}
\frac{d\bm{z}_t}{dt} = \bm{\epsilon}_{\bm{\theta}}(\bm{z}_t, t).
	\end{aligned}
\end{equation}

Similar to the channel denoising DM in \cite{10480348}, wireless channel denoising task is a subprocess of whole diffusion reverse process. The denoising start point $t_m$ should be determined by $\mathop{\arg\min}_{t_m}\left | \sigma^2 - t_m^2 \right |$ with known $\sigma^2$ and $m$ denotes the utilized denoising steps of pretrained LDM. Consequently, The selection of hyperparameters $N$ and $T$ should consider the worst-case SNRs to make the channel denoising objective a sub-term of the DM training objective. Ultimately, the transmitted latent vector $\bm{z}_R$ is given by $\bm{H}^{-1}_z\bm{z}_0$. Nonetheless, in the wireless communication scenarios with large noise variance $\sigma^2$, $m\gg 1 $ according to the designed discrete schedule $\left \{ t_n\right \}^{n=N}_{n=0}$. As a result, LDM will execute $m$ times of noise predictions $\bm{\epsilon}_{\bm{\theta}}(\bm{z}_t,t)$, i.e., the number of function evaluations (NFE) will reach $m$. Unfortunately, the varying fading wireless channels with uncertain SNRs bring significant uncertainties to the computational complexity of LDM, which undermines the possibility of implementing real-time SemCom.

\subsection{End-to-End Consistency Distillation} \label{EECD}
The multi-step reverse sampling process of DMs brings the disadvantage of slow data generation speed. To overcome it, methods based on denoising diffusion implicit model (DDIM) subsequence sampling \cite{song2020denoising}, optimal reverse variances \cite{10542391}, LDMs \cite{rombach2022high}, SDE/ODE solvers \cite{zhou2023fast}, and knowledge distillation \cite{song2023consistency} have been proposed to optimize or accelerate the sampling process. In detail, the LDMs can significantly reduce the demensionality of input data, and some distillation based approaches only require a few steps or even one step to evaluate the output data without generation quality issues. Among these acceleration methods, the consistency model \cite{song2023consistency}, as one of the distillation approaches, defines the consistency function $\bm{f}: (\bm{z}_t,t) \mapsto \bm{z}_\varepsilon$ given a forward trajectory $\left \{ \bm{z}_t \right \}_{t \in [\varepsilon ,T]}$, where $\varepsilon = t_1 \approx 0$. The consistency function assumes that for the input data on the same forward trajectory, the output of the neural network parameterized function points to the same generated data, which is given by 
\begin{equation}
	\begin{aligned}
\bm{f}_{\hat{\bm{\theta}}}(\bm{z}_t,t)=
\begin{cases}
\bm{z}_t & {t=\varepsilon}\\
\bm{F}_{\hat{\bm{\theta}}}(\bm{z}_t,t)& {t \in (\varepsilon, T]},
\end{cases}
	\end{aligned}
\end{equation} 
where $\hat{\bm{\theta}}$ is the neural network parameters of consistency model. By observing function $\bm{F}_{\hat{\bm{\theta}}}(\bm{z}_t,t)$, it can be implemented by directly training a neural network to map noisy data $\left \{ \bm{z}_t \right \}_{t \in (\varepsilon, T]}$ to $\bm{z}_{\varepsilon}$.  Accordingly, the consistency function $\bm{f}_{\hat{\bm{\theta}}}(\bm{z}_t,t)$ can be obtained based on EDM architecture by distilling the pretrained original LDM $\bm{\epsilon}_{\bm{\theta}}(\bm{z}_t,t)$.

\begin{figure}[!t]
	\vspace{-0.1cm}
	\centerline{\includegraphics[width=0.5\textwidth]{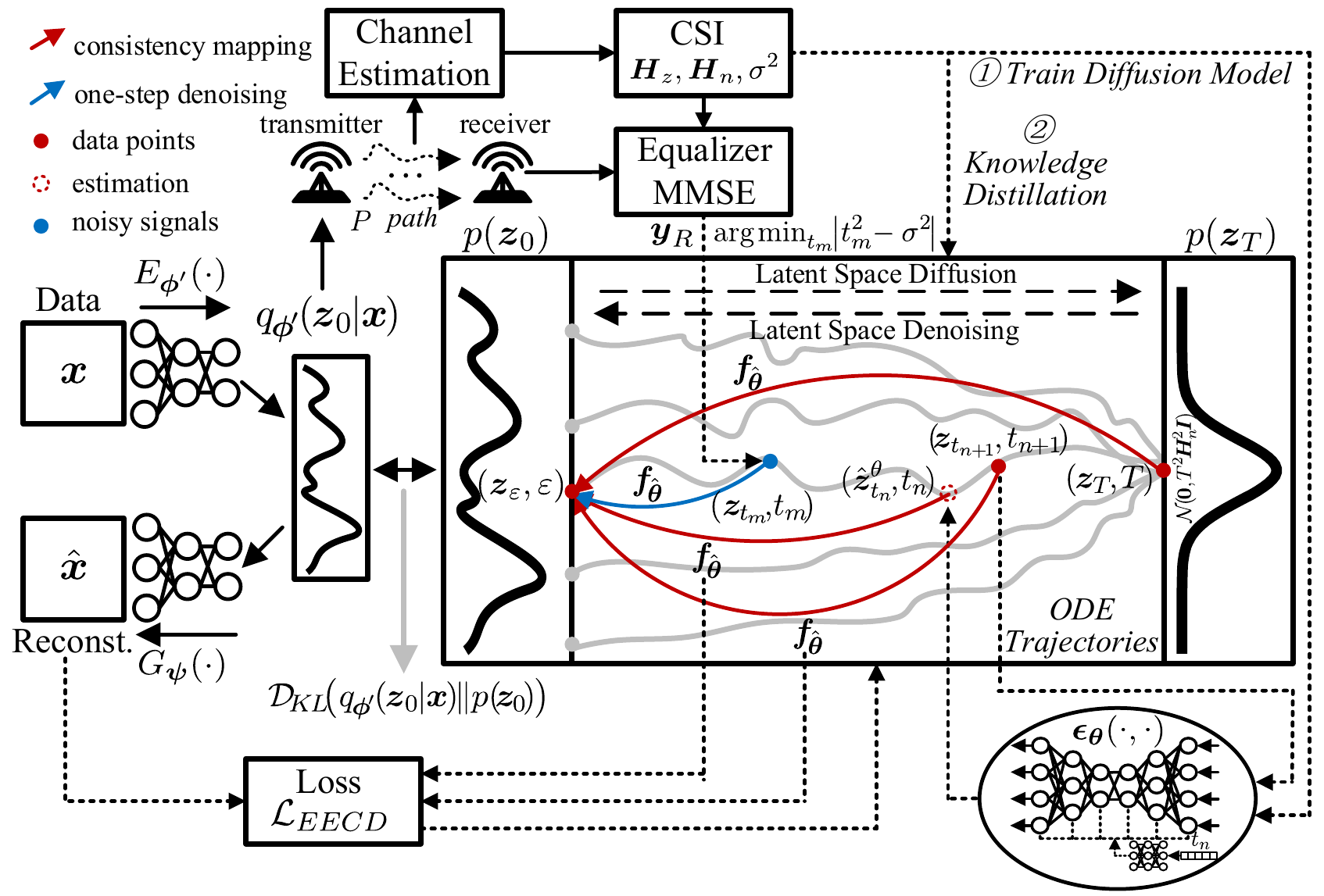}}
	\caption{In the proposed SemCom model, data is mapped into latent space via robust encoder $q_{\bm{\phi}'}(\bm{z}_0|\bm{x})$. Then, EECD maps noisy received signals to denoised latent vector ($\bm{z}_{t_m} \rightarrow \bm{z}_{\varepsilon }$) and decoder will generate data with desired semantic meaning by $p_{\bm{\psi}}(\bm{x}|\bm{z}_{\varepsilon })$.}
	\label{Task3}
	\vspace{-0.1cm}
\end{figure}

\begin{algorithm}[!t]  	\label{trainingEECD}
	\small
	\caption{Training algorithm of EECD}
		\LinesNumbered
		\KwIn{Dataset $q(\bm{x})$, initial model parameter $\hat{\bm{\theta}}$, robust encoder $E_{\bm{\phi}}(\cdot)$, generator $G_{\bm{\psi}}(\cdot)$, pretrained latent diffusion model $\bm{\epsilon}_{\bm{\theta}}(\cdot,\cdot)$, distance metric $\bm{d}(\cdot, \cdot)$, learning rate $\eta$, decay rate $\mu$, time schedule $\left \{ t_n\right \}^{t=N}_{t=1}$, and channel state information $\bm{H}_z,\bm{H}_n$ }
		\KwOut{The trained one-step end-to-end consistency model}
		\textbf{Initialize} $\hat{\bm{\theta}}^{-} \leftarrow \hat{\bm{\theta}}$\;
        \Repeat{Converged}{
        Sample $\bm{x}\sim q(\bm{x})$ and $n \sim \mathcal{U}[1,N-1]$  \;
		Compute $\bm{z} \leftarrow E_{\bm{\phi}}(\bm{x})$ and transmitted $\bm{z}_R$\;
		Sample $\bm{z}_{t_{n+1}} \sim \mathcal{N}\left ( \bm{z}_{t_{n+1}}; \bm{H}_z\bm{z}_R, t^2_{n+1}\bm{H}^2_n\bm{I}\right )$\;
		Compute $\tilde{\bm{z}}_{t_n}^{\bm{\theta}} \leftarrow \bm{z}_{t_{n+1}} - \bm{\epsilon}_{\bm{\theta}}(\bm{z}_{t_{n+1}}, t_{n+1})(t_{n+1}-t_n)$\;
		Estimate $\bm{z}_{t_n}$ by $\hat{\bm{z}}_{t_n}^{\bm{\theta}} \leftarrow \bm{z}_{t_{n+1}} -\frac{1}{2} \Big [\bm{\epsilon}_{\bm{\theta}}(\tilde{\bm{z}}^{\bm{\theta}}_{t_{n}}, t_{n}) +\bm{\epsilon}_{\bm{\theta}}(\bm{z}_{t_{n+1}}, t_{n+1})  \Big ] (t_{n+1}-t_n)$\;
		Compute $\mathcal{L}_{EECD}\left ( \hat{\bm{\theta}}, \hat{\bm{\theta}}^{-} | \bm{\theta}, \bm{\psi} \right )$ by Eq. \eqref{EECDloss}\;
		Update $\hat{\bm{\theta}} \leftarrow \hat{\bm{\theta}} - \eta \nabla_{\hat{\bm{\theta}}}\mathcal{L}_{EECD}\left ( \hat{\bm{\theta}}, \hat{\bm{\theta}}^{-} | \bm{\theta}, \bm{\psi} \right ) $\;
		Update $\hat{\bm{\theta}}^{-} \leftarrow \textrm{stopgrad}\left ( \mu\hat{\bm{\theta}}^{-}+ (1-\mu)\hat{\bm{\theta}}  \right )$\;
		}
		\textbf{Return} End-to-end distillated consistency model $\bm{f}_{\hat{\bm{\theta}}}(\cdot, \cdot)$
\end{algorithm}

Assume that the time schedule of the sampling process is $\varepsilon = t_1<T_2<\cdots <t_N =T$, the Euler solver is adopted for reverse process evaluation. As a result, at $t=t_{n+1}$, Eq. \eqref{PF-ODE2} can be transformed into 
\begin{equation} \label{Eulersolver}
	\begin{aligned}
& \left . \frac{d\bm{z}_t}{dt}\right |_{t=t_{n+1}} = \bm{\epsilon}_{\bm{\theta}}(\bm{z}_{t_{n+1}},t_{n+1}) \approx \frac{\bm{z}_{t_{n+1}}- \bm{z}_{t_n}}{t_{n+1}-t_n}  \\
\Leftrightarrow & \tilde{\bm{z}}^{\bm{\theta}}_{t_n} \approx \bm{z}_{t_{n+1}}- \bm{\epsilon}_{\bm{\theta}}(\bm{z}_{t_{n+1}},t_{n+1})\left ( t_{n+1}-t_n \right ),
	\end{aligned}
\end{equation} 
where $\tilde{\bm{z}}^{\bm{\theta}}_{t_n}$ denotes the predicted data point at $t=t_{n}$. This is also called denoising diffusion implicit model (DDIM) \cite{song2020denoising} and every step's NFE equals to 1. However, the actual values of the difference $\frac{\bm{z}_{t_{n+1}}-\bm{z}_{t_n}}{t_{n+1}-t_n}$ are closer to the derivatives between $\bm{z}_{t_{n+1}}$ and $\bm{z}_{t_n}$, rather than the dervative at $\bm{z}_{t_{n+1}}$. To this end, the Heun solver in EDM is adopted \cite{karras2022elucidating}, which is denoted by
\begin{equation} \label{Heunsolver}
	\begin{aligned}
& \frac{\bm{z}_{t_{n+1}}- \bm{z}_{t_n}}{t_{n+1}-t_n}   \approx  \frac{1}{2} \left ( \bm{\epsilon}_{\bm{\theta}}(\bm{z}_{t_{n}},t_{n})  + \bm{\epsilon}_{\bm{\theta}}(\bm{z}_{t_{n+1}},t_{n+1})  \right )  \Leftrightarrow \\
 & \hat{\bm{z}}^{\bm{\theta}}_{t_n}  \approx \bm{z}_{t_{n+1}}- \frac{1}{2} \left ( \bm{\epsilon}_{\bm{\theta}}(\bm{z}_{t_{n}},t_{n})+ \bm{\epsilon}_{\bm{\theta}}(\bm{z}_{t_{n+1}},t_{n+1}) \right ) \left ( t_{n+1}-t_n \right ),
	\end{aligned}
\end{equation} 
where $\bm{z}_{t_n}$ on the right side of Eq. \eqref{Heunsolver} can be approximated by $\tilde{\bm{z}}^{\bm{\theta}}_{t_n}$ in Eq. \eqref{Eulersolver}. Consequently, the estimation of $\bm{z}_{t_n}$ is given by 
\begin{equation} 
	\begin{aligned}
\hat{\bm{z}}^{\bm{\theta}}_{t_n}  \approx \bm{z}_{t_{n+1}}- \frac{1}{2} \left ( \bm{\epsilon}_{\bm{\theta}}(\tilde{\bm{z}}^{\bm{\theta}}_{t_{n}},t_{n})+ \bm{\epsilon}_{\bm{\theta}}(\bm{z}_{t_{n+1}},t_{n+1}) \right ) \left ( t_{n+1}-t_n \right ),
	\end{aligned}
\end{equation} 
where the NFE equals to 2. According to the definition of the consistency function, the function $\bm{f}_{\hat{\bm{\theta}}}(\bm{z}_t,t)$ should have the same output for adjacent data points $(\bm{z}_{t_{n+1}}, t_{n+1})$ and $(\hat{\bm{z}}^{\bm{\theta}}_{t_n}, t_n)$ on the same reverse trajectory, i.e., the loss of the consistency model is
\begin{equation} \label{CDloss}
	\begin{aligned}
\mathcal{L}_{CD}\left ( \hat{\bm{\theta}}, \hat{\bm{\theta}}^{-} | \bm{\theta} \right ) = \mathbb{E}_q \left [ \bm{d} \left ( \bm{f}_{\hat{\bm{\theta}}}(\bm{z}_{t_{n+1}}, t_{n+1}),  \bm{f}_{\hat{\bm{\theta}}^{-}}(\hat{\bm{z}}^{\bm{\theta}}_{t_n}, t_n)\right )\right ],
	\end{aligned}
\end{equation} 
where $\hat{\bm{\theta}}^{-}$ denotes the running average of the past values of $\hat{\bm{\theta}}$ during optimization. 

Nonetheless, the goal of wireless SemCom is to accurately reconstruct the transmitted data in real-time manner in the receiver side. Inspired by that, the consistency distillation (CD) loss in Eq. \eqref{CDloss} can be changed to the loss of EECD, which is given by
\begin{equation} \label{EECDloss}
	\begin{aligned}
& \mathcal{L}_{EECD}\left ( \hat{\bm{\theta}}, \hat{\bm{\theta}}^{-} | \bm{\theta}, \bm{\psi} \right ) \\
& = \mathbb{E}_q \left [ \bm{d} \left ( G_{\bm{\psi}}\left ( \bm{f}_{\hat{\bm{\theta}}}(\bm{z}_{t_{n+1}}, t_{n+1}) \right ),  G_{\bm{\psi}} \left (\bm{f}_{\hat{\bm{\theta}}^{-}}(\hat{\bm{z}}^{\bm{\theta}}_{t_n}, t_n) \right ) \right )\right ],
	\end{aligned}
\end{equation} 
where $\bm{d}(\cdot, \cdot)$ is denoted by Euclidean distance for non-images datasets, and structural similarity index measure (SSIM) or learned perceptual image path similarity (LPIPS) \cite{zhang2018unreasonable} based distance for image datasets.

The distillation training process of EECD is illustrated in Algorithm \ref{trainingEECD}. As depicted in Fig. \ref{Task3}, the student model $\bm{f}_{\hat{\bm{\theta}}}(\cdot, \cdot)$ updates its parameters $\hat{\bm{\theta}}$ and $\hat{\bm{\theta}}^-$ through gradient descent by minimizing the perceptual loss $\mathcal{L}_{EECD}$ between the decoded data $(\bm{x}_{t_{n+1}}, t_{n+1})$ and $(\hat{\bm{x}}^{\bm{\theta}}_{t_n}, t_n)$ corresponding to the original latent space data point $(\bm{z}_{t_{n+1}}, t_{n+1})$ and the next point $(\hat{\bm{z}}^{\bm{\theta}}_{t_n}, t_n)$ in the reverse diffusion process predicted by the diffusion (teacher) model. This process ensures the consistency between denoised signals by direct mapping to $\bm{z}_\varepsilon$ along the same ODE trajectory. Additionally, in the sampling phase, the pretrained latent consistency model can flexibly enhance the perceptual quality of reconstruction by resampling $s-1$ times based on the subsequence $\bm{\tau} = [\tau_1, \tau_2, \cdots, \tau_s]$ of length $s$, where $\tau_1 = m$. Consequently, the real-time channel denoising and data reconstruction process based on EECD model is given in Algorithm \ref{samplingEECD}. The advantages and contributions of the proposed LDM approach are further elaborated as follows:

\begin{itemize}[leftmargin=10pt]
\item VE, SDE, and PF-ODE are utilized to model the LDM and wireless channel denoising processes. The advantage of this novel approach lies in its clearer interpretation of physical channels, making it more intuitive and capable of accommodating various channel conditions. 
\item The training of the original LDM cannot optimize the generation of latent space together with the decoder $G_{\bm{\psi}}(\cdot)$. However, the proposed end-to-end consistency loss allows the training objective to no longer be limited to mapping received noisy signal  $\bm{y}_R$ after equalization to the denoised latent space $\hat{\bm{z}}_0$, but directly measures the distance of adjacent data points on the same trajectory.
\item The EECD based loss effectively eliminates the limitation of only being able to calculate the Euclidean distance between two latent bottlenecks and multi-step diffusion processes. Consequently, the latent consistency model can directly utilize more superior semantic metrics such as LPIPS to enhance perceptual quality.

\end{itemize}

\begin{algorithm}[!t]  	\label{samplingEECD}
	\small
	\caption{Sampling of channel denoising EECD}
		\LinesNumbered
		\KwIn{Transmitted data $\bm{x}$, robust encoder $E_{\bm{\phi}'}(\cdot)$, generator $G_{\bm{\psi}}(\cdot)$, distillated end-to-end consistency model $\bm{f}_{\hat{\bm{\theta}}}(\cdot,\cdot)$, subsequence length $s$, and channel state information $\bm{H}_z, \bm{H}_n, \sigma$ }
		\KwOut{Reconstructed data $\hat{\bm{x}}$ at the receiver}
		Compute the encoded latent space $\bm{z} \leftarrow E_{\bm{\phi}'}(\bm{x})$\;
		Transmit real-valued $\bm{z}_R$ through noisy wireless channel\;
		Compute MMSE equalization $\bm{y}_R \leftarrow \bm{H}_z\bm{z}_R + \bm{H}_n\sigma \bm{\epsilon}$ and $\bm{\epsilon}\sim \mathcal{N}(\bm{0},\bm{I})$\;
		Estimate $t_m$ by ${\arg\min}_{t_m}\left | t_m^2-\sigma^2\right |$\;
		Compute denoised estimation $\hat{\bm{z}}_{\varepsilon} \leftarrow \bm{f}_{\hat{\bm{\theta}}}(\bm{y}_R, t_m)$\;
		\If{$s>1$}{
			Determine subsequence $\bm{\tau} = [\tau_1, \tau_2, \cdots, \tau_s]$\;
			\For{$i = 2$ \textup{\textbf{to}} $s$}{
                Sample $\bm{z}_{t_{\tau_i}} \sim \mathcal{N} (\bm{z}_{t_{\tau_i}}; \hat{\bm{z}}_{\varepsilon}, t_{\tau_i}^2\bm{H}^2_{n} \bm{I})$\;
				Compute $\hat{\bm{z}}_{\varepsilon} \leftarrow \bm{f}_{\hat{\bm{\theta}}}(\bm{z}_{t_{\tau_i}}, t_{\tau_i})$\;
			}
		}
		Compute denoised data $\hat{\bm{z}}_R \leftarrow \bm{H}_z^{-1} \hat{\bm{z}}_{\varepsilon}$ and decoded $\hat{\bm{z}}$\;
		\textbf{Return} the recovered data $\hat{\bm{x}} \leftarrow G_{\bm{\psi}}(\hat{\bm{z}})$
\end{algorithm}

\section{Numerical Experiments}  
\label{sec:NE}
\subsection{Experimental Setup} \label{expersetup}

\begin{figure*}[!t]
	\vspace{-0.1cm}
	\centerline{\includegraphics[width=0.96\textwidth]{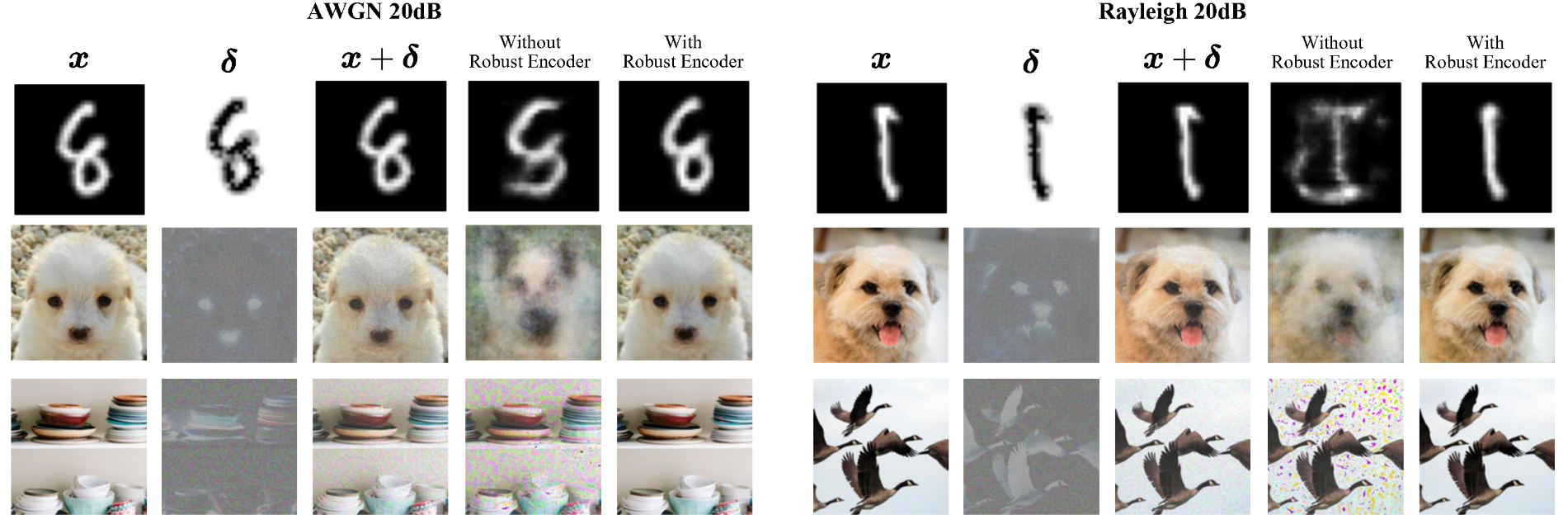}}
	\caption{ Some typical decoded images without/with robust encoder under AWGN and Rayleigh channel. The SNR is 20dB. }
	\label{Fig1}
	\vspace{-0.4cm}
\end{figure*}

\textbf{\textit{1) Dataset:}} MNIST handwritten digit image dataset is initially considered for evaluating the proposed SemCom system, containing 60,000 images for training and 10,000 for testing. Additionally, to validate the performance of \textit{out-of-domain} adaptation, the Fashion-MNIST (F-MNIST) dataset is also employed in the evaluation, comprising images of various types of clothing and accessories, with an identical distribution of 60,000 training and 10,000 testing images. The resolution for both MNIST and F-MNIST datasets are uniformly resized to 32×32. Furthermore, the animal face high quality (AFHQ) dataset \cite{choi2020stargan} is also selected to verify the effectiveness of the proposed method, including a total of 15,000 RBG images across three categories: dogs, cats, and wild animals, where 4,500 images of dogs are used for training and the remaining 500 dog images, along with 500 cat images, are used to test the proposed method, with the resolution resized to 192×192. Lastly, the DIV2K high-quality RGB image dataset \cite{Timofte_2018_CVPR_Workshops} is also considered for SemCom tasks, encompassing 800 diverse training images, 100 validation images, and 100 test images, with the resolution resized to 256×256.

\textbf{\textit{2) Baseline Method:}} Four distinct implementations of communication systems are utilized for comparison to demonstrate the superiority of the proposed SemCom system. The first is a combination of the state-of-the-art traditional image compression method JPEG2000 \cite{christopoulos2000jpeg2000} with the error correction technique LDPC \cite{chen2005reduced}, denoted as JPEG2000+LDPC. The second is the widely recognized CNN-based deep JSCC method \cite{8723589}, where joint source-channel training effectively mitigates the adverse effects of unreliable channels. The third is the multi-step VE-based LDM (denoted as VE-LDM), and the fourth is accelerated DDIM, which involves 2-step sampling \cite{10542391}, both belonging to the DM-aided approaches.

\textbf{\textit{3) Performance Metrics:}} The metrics for model evaluation can broadly be categorized into two types. The first category encompasses traditional image reconstruction metrics for bit/symbol accuracy, including measures such as mean squared error (MSE)$\downarrow$ and peak SNR (PSNR)$\uparrow$, where $\uparrow$ indicates that a higher value represents better performance, while $\downarrow$ indicates the opposite. The second category consists of semantic or human-perceptual metrics that warrant increased attention within the context of SemCom. For image transmission, this includes the SSIM and multi-scale SSIM (MS-SSIM)$\uparrow$ \cite{dai2022nonlinear} and the pretrained VGG-based LPIPS$\downarrow$ \cite{zhang2018unreasonable}.

\textbf{\textit{4) CSI Condition:}} For the wireless channels, three distinct channels are taken into consideration, including the AWGN channel ($K = \infty $), the Rayleigh channel ($K=0$), and the Rician channel ($K=1$). Regarding the noise level of the channels, noise with SNRs ranging from 0 dB to 20 dB is contemplated for testing the performance of various methods under different SNR conditions. The channel bandwidth ratio (CBR) is defined as $\textrm{CBR}=k/(H\times W\times C)$, where $H$, $W$, and $C$ are the height, width (resolution) and colour channel of images, and usually $H$=$W$. CBR is also an exceedingly crucial metric in SemCom, defining the demand for communication resources \cite{dai2022nonlinear, 10480348}. For this reason, CBRs from 0.01 to 0.05 are implemented on DL models trained by AFHQ and DIV2K datasets, while for the MNIST dataset, due to its low resolution, only the DL with 1/16 CBR and the JPEG2000+LDPC with 1/3 CBR have been realized.

\textbf{\textit{5) Simulation Environment and Hyperparameters:}} The simulations are conducted using Python 3.8.19 and CUDA-accelerated PyTorch 2.3.0 on a computer equipped with an i5-13600KF CPU operating at 3.50GHz, 32 GB of RAM, and an NVIDIA GeForce RTX 4070 GPU. In the encoder and decoder parts, $G_{\bm{\psi}(\cdot)}$ and $E_{\bm{\phi}'}(\cdot)$ each contain 7 transposed convolution layers and 6 convolution layers. The training for the wireless channel denoising task can follow a shorter time schedule. Consequently, the total length of the forward process for the LDM is set to $N=100$, with the variance starting point at $t_1 = \varepsilon = 0.002$ and the endpoint at $t_N = T = 2$, and $\rho=7.0$. Furthermore, the learning rate during training is established at 1e-4, with an initial decay rate of 0.95 and a decay rate of 0.99993 for the student model.

\textbf{\textit{6) Training, Deployment and Testing:}} During the training and deployment phases, first, the convolutional WGAN and VAE are trained sequentially or the Algorithm \ref{trainingVAEWGAN} is jointly trained and deployed over rate-limited channels; then, the parameters of the trained convolutional VAE will be fine-tuned into a robust encoder in a self-supervised learning manner following the steps of Algorithm \ref{trainingrobustencoder}; finally, the learning of the parameters of the diffusion model is conducted end-to-end according to the denoising EECD strategy in Algorithm \ref{trainingEECD}. In the testing phase, LDM denoises the received equalized signals according to Algorithm \ref{samplingEECD}, and when the reconstruction error exceeds a threshold, the Algorithm \ref{trainingadaptor} will be activated to adjust latent vector $\bm{z}$ for the low-precision data.

\subsection{Robustness to Data Inaccuracies}

As stated in Subsection \ref{robustencoder}, the encoder parameters can be updated via augmented learning based on the obtained semantic errors $\bm{\delta}$. For the MNIST, AFHQ, and DIV2K datasets, pretrained encoders are updated with a learning process at an error level of $\left \| \bm{\delta} \right \|_p/H = 0.3$. Following the update, several prototypical image datasets are employed to test the robust encoder's efficacy in effectively countering data inaccuracies. Fig. \ref{Fig1} illustrates the impact of semantic errors with levels of 0.5 and 0.4 superimposed on the original data under AWGN channel and Rayleigh channel at SNR of 20 dB, respectively. It is readily observed that with the proposed robust encoder, the source data with added semantic errors still bear minimal semantic differences from the original data to the human visual perception. However, when the original SemCom system, without a robust encoder, transmits this contaminated data, the decoded output can result in significant semantic errors, as shown in Fig. \ref{Fig1} for the example images in MNIST and AFHQ datasets, and might also lead to extensive artifacts in reconstructed images as seen for the DIV2K dataset. Fortunately, the introduction of the robust encoder successfully overcomes semantic ambiguities that may arise from the contaminated data due to cyber attacks or other types of outliers, ensuring that the decoded data at the receiver still carries the correct semantic information.

\begin{table*}[!h]\centering \footnotesize
	\caption{Robustness of semantic communication system under different levels of semantic errors and Gaussian nioses (without robust encoder/with robust encoder), where CBR is fixed at 0.0208 and CSI is varying}
	\label{robust_encoder_results_1}
	\renewcommand{\arraystretch}{1.1}
	\setlength\tabcolsep{0.5em}
	\begin{tabular}{ccccccccccc}
	\toprule
	\toprule
	\multicolumn{2}{c}{Error/Noise Metric}     & \multicolumn{5}{c}{$\left \| \bm{\delta} \right \|_p/H$} & \multicolumn{4}{c}{ SNR (dB)   } \\
	\cmidrule(lr){1-2} \cmidrule(lr){3-7} \cmidrule(lr){8-11}
	
	\multicolumn{2}{c}{Error/Noise Level}         &  0.1 & 0.2 & 0.3 & 0.4 & 0.5 &  5  & 7.5 & 10 & 12.5   \\
	\midrule
	 \multirow{3}{*}{MNIST}     & PSNR (dB)$\uparrow$   &  16.54/18.50     &  11.88/18.52    &   8.33/\textbf{18.60}  & 5.96/16.36    &  5.16/12.42    & 6.17/8.16 &  7.52/11.35     &  10.05/14.91    &   12.30/\textbf{17.10}               \\
	& SSIM (dB)$\uparrow$   &  10.65/13.32     &  5.72/\textbf{13.56}    &  2.77/13.19   &  1.18/10.56   &   0.76/6.79   &   1.40/3.52  &  2.21/5.15     &  4.01/8.60    &  5.77/\textbf{11.18}                     \\
			     & MSE$\downarrow$   &  0.022/0.014     &  0.065/0.014    &  0.147/\textbf{0.013}   &  0.253/0.023   &   0.304/0.057   &   0.241/0.153  &  0.177/0.073     &  0.099/0.032    &  0.059/\textbf{0.019}                 \\				 
                \midrule
	\multirow{3}{*}{AFHQ} & PSNR (dB)$\uparrow$     & 22.31/\textbf{22.51}      &  18.94/22.14    &   15.34/21.58  &  12.82/21.38   &  9.72/20.58       & 13.56/15.24      &  18.45/19.19    &   21.20/21.89  &  \textbf{22.32}/21.19       \\
			        & MS-SSIM (dB)$\uparrow$     &  19.82/\textbf{20.68}     &  13.75/20.65    &  9.34/20.28   & 6.56/19.11   &   3.81/17.76     &  6.76/9.86     &  13.42/14.27    &  17.92/18.44   & 20.27/\textbf{20.40}     \\
					& LPIPS$\downarrow$     &  0.160/\textbf{0.152}     &  0.211/0.157    &  0.302/0.158   &  0.410/0.172    &   0.531/0.180     &  0.475/0.348     &  0.246/0.226    &  0.175/0.172   & 0.154/\textbf{0.151}      \\
				\midrule
	\multirow{3}{*}{DIV2K} & PSNR (dB)$\uparrow$     & 23.60/\textbf{23.71}      &  18.62/23.17    &   14.55/22.70  &  10.37/22.01   &  8.99/21.65      & 11.87/14.27      &  16.54/18.66    &   20.87/21.24  &  22.20/\textbf{22.41}       \\
				& MS-SSIM (dB)$\uparrow$     &  16.19/\textbf{16.49}     &  10.20/16.28    &  8.22/16.01   & 6.01/15.88   &   4.94/15.19     &  5.66/8.39     &  11.09/12.73   &  13.68/15.11   & 15.35/\textbf{16.22}      \\
				& LPIPS$\downarrow$     &  0.122/\textbf{0.121}     &  0.208/0.129    &  0.325/0.133   & 0.442/0.147    &   0.560/0.157     &  0.512/0.297     &  0.261/0.237    &  0.184/0.160   & 0.131/\textbf{0.128}      \\				
	\bottomrule
	\bottomrule
	\end{tabular}
	\vspace{-0.4cm}
	\end{table*}

To maintain generality, the results of multiple performance tests for various outlier types and levels across different datasets are documented in Table \ref{robust_encoder_results_1}. Specifically, the CBR is fixed at approximately 0.02 and the test CSI conditions vary in accordance with Subsection \ref{expersetup}. It is not difficult to observe that the robust encoder, despite only being updated at a semantic error level of 0.3, still maintains robustness compared to the original encoder under other semantic error levels and low-SNR noise contamination, thereby enhancing the quality of decoded data when source data is subject to semantic errors or noises. Furthermore, evaluation metrics such as PSNR/SSIM/MS-SSIM can be improved by several times or even an order of magnitude, while MSE/LPIPS can be reduced by several times or even by an order of magnitude.

\subsection{Out-of-Domain Adaptation}

\begin{figure}[!h]
	\vspace{-0.1cm}
	\centerline{\includegraphics[width=0.48\textwidth]{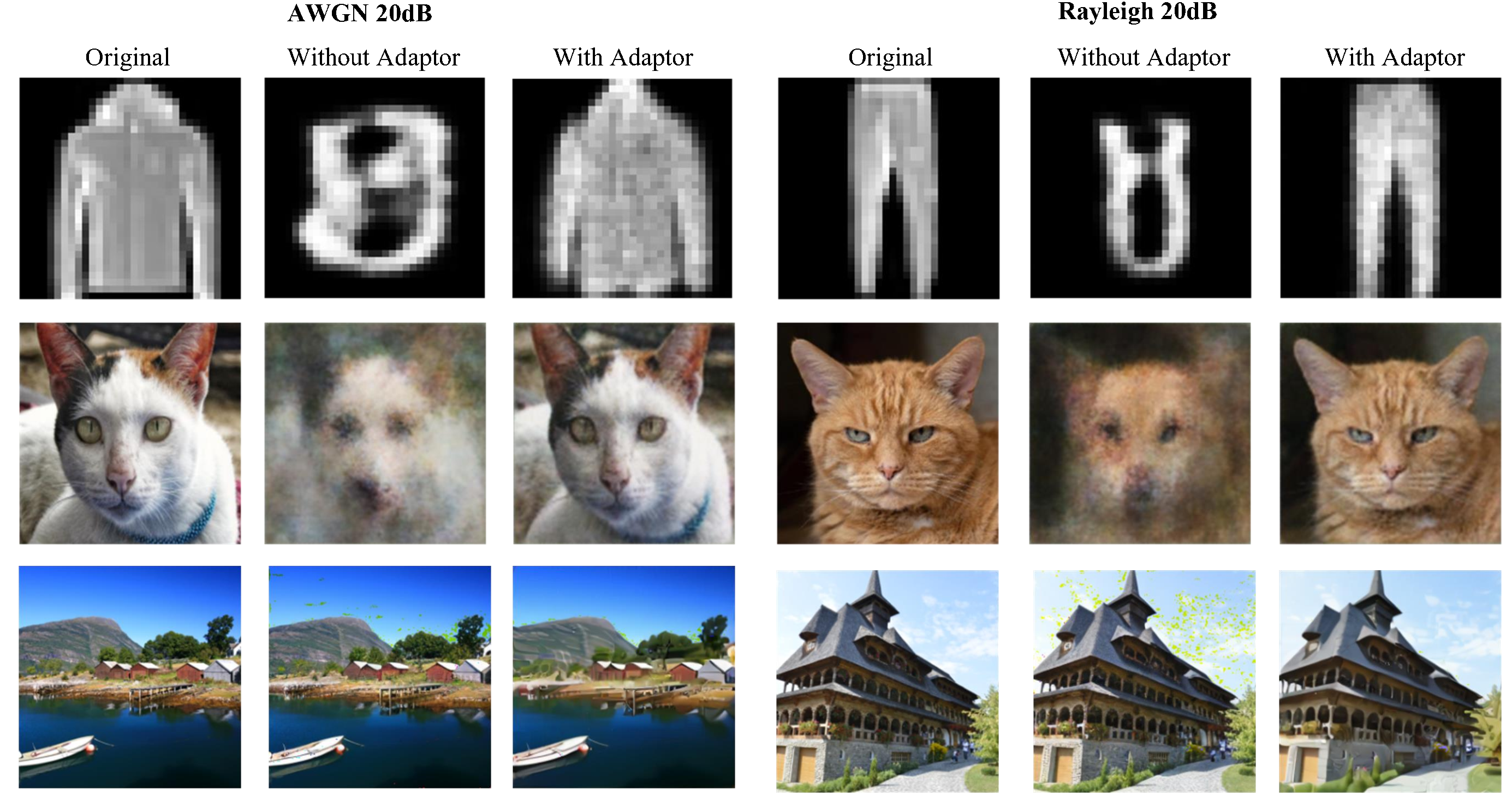}}
	\caption{Some typical decoded images without/with adaptor in AWGN and Rayleigh channel. The SNR is 20 dB.
	}
	\label{Fig2}
	\vspace{-0.1cm}
\end{figure}

As described in Subsection \ref{ofdls}, the proposed SemCom system employs a lightweight, single-layer adapter at the transmitter for rapid one-shot learning and transforms the latent space at the receiver, thereby enabling the DL-based SemCom system to adapt to \textit{out-of-domain} data or enhance decoding quality. Specifically, subsets of clothing images from the F-MNIST dataset and cat images from the AFHQ dataset were utilized to validate the efficacy of the adapter. As illustrated in Fig. \ref{Fig2}, without the adapter enabled, a SemCom system pretrained with a particular type of data would decode data at the receiver that more closely resembles that specific type of semantic information, leading to severe semantic ambiguity. However, the adapter situated before the generator can swiftly overcome this shortcoming, producing data that is essentially consistent with the original semantics of the transmitted data. Additionally, the original training DL model underperformed on certain test data from the DIV2K dataset, with decoded data exhibiting partial errors. The adapter also enhances communication quality in such instances, eliminating artifacts in the images.

\begin{figure}[!h]
	\vspace{-0.1cm}
	\centerline{\includegraphics[width=0.42\textwidth]{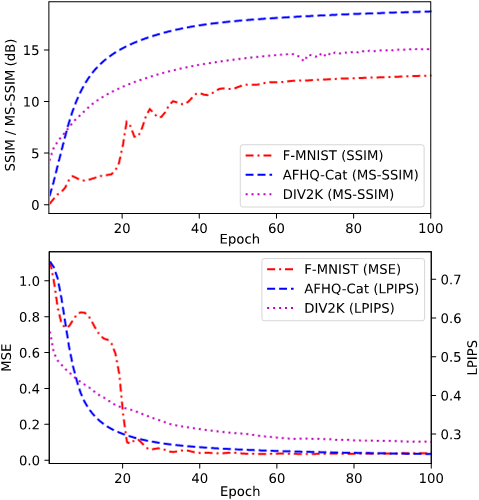}}
	\caption{The performance metrics' curves when doing one-shot learning to update the parameters of $\bm{g}_{\bm{\omega}}(\cdot)$.
	}
	\label{Fig4}
	\vspace{-0.1cm}
\end{figure}

The evolution of performance metrics during the one-shot learning process for the three datasets is depicted in Fig. \ref{Fig4}. Evidently, after approximately only 20 epochs, the metrics of decoded data with adapters can be swiftly ameliorated to ideal values, thereby diminishing semantic ambiguities. To ensure generality, numerical experiments are also conducted to corroborate the effectiveness of the proposed adaptive strategy in enhancing SemCom performance and mitigating semantic ambiguity, with the results presented in Table \ref{domain_adaptation_results}. Notably, as evidenced by the semantic evaluation metrics SSIM/MS-SSIM and LPIPS, the incorporation of adapters substantially augments the receiver's \textit{out-of-domain} adaptation and reconstruction capabilities under certain constraints on image categories, preventing the emergence of semantic ambiguities.	

\begin{table}[!h]\centering \small
	\caption{Improvement in adapation and reconstruction performance for different types of data (without adaptor/with adaptor)}
	\label{domain_adaptation_results}
	\renewcommand{\arraystretch}{1.1}
	\setlength\tabcolsep{0.2em}
	\begin{tabular}{cccc}
	\toprule
	\toprule
	\diagbox{Dataset}{Metrics}    & PSNR (dB)$\uparrow$ & SSIM/MS-SSIM (dB)$\uparrow$ & MSE/LPIPS$\downarrow$ \\
    \midrule
    F-MNIST & 6.16/13.82 & 0.48/8.90 & 0.313/0.049 \\
	AFHQ-Cat & 9.93/19.56 & 3.09/16.59 & 0.655/0.232\\
    DIV2K & 17.63/28.67 & 10.51/16.30 & 0.288/0.175 \\
	\bottomrule
	\bottomrule
	\end{tabular}
	\vspace{-0.0cm}
	\end{table}

\begin{figure*}[!h]
		\vspace{-0.1cm}
		\centerline{\includegraphics[width=0.92\textwidth]{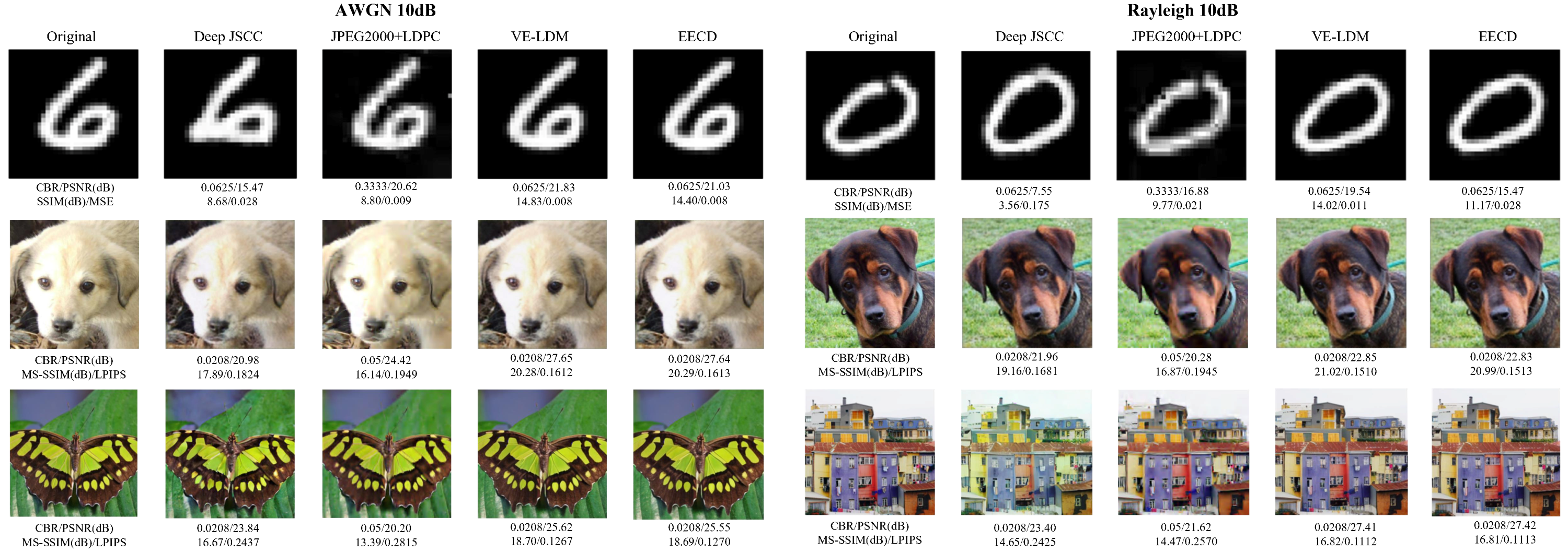}}
		\caption{Some typical decoded images with received signals denoised by different approaches under AWGN and Rayleigh channel. The SNR is 10 dB. }
		\label{Fig5}
		\vspace{-0.2cm}
\end{figure*}

\begin{figure*}[!h]
	\centering   
	\subfigure 
	{
        
		\includegraphics[width=0.95\linewidth]{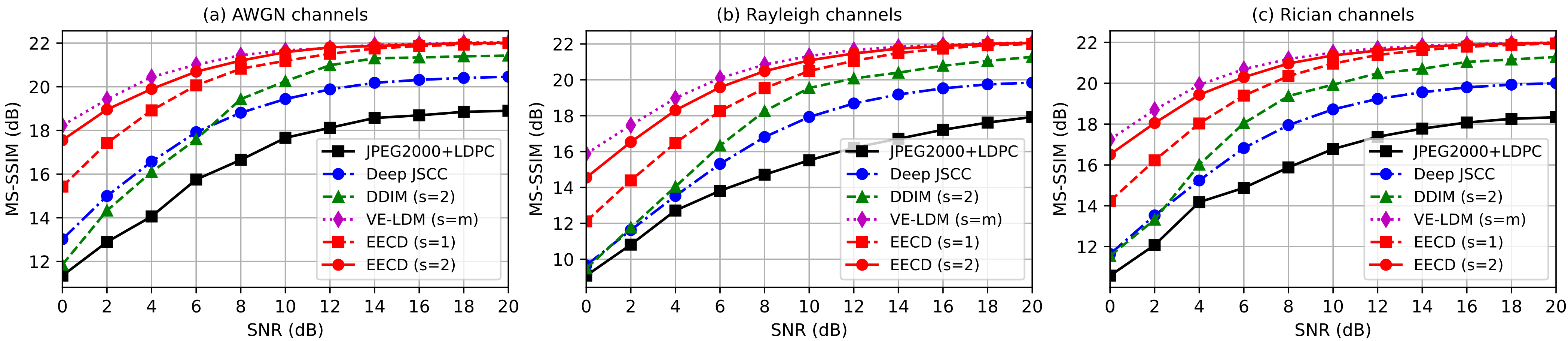}
	}

	\vspace{-0.8em}

	\subfigure
	{
		\includegraphics[width=0.95\linewidth]{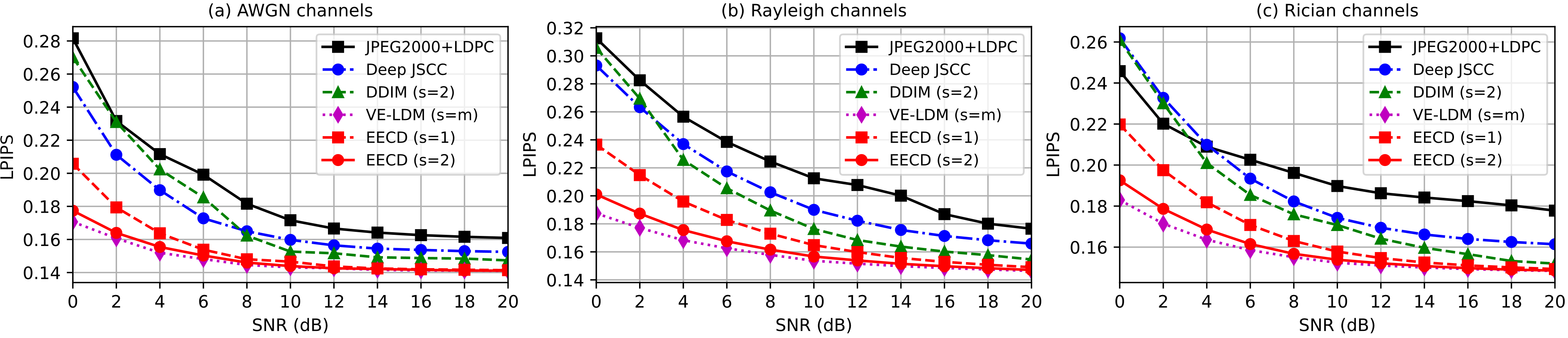}

	}
	\caption{Semantic performance metrics of JPEG2000+LDPC, Deep JSCC, DDIM, VE-LDM, and EECD methods under different SNRs and channel states within AFHQ dataset.}
	\vspace{-0.5cm}
	\label{Fig6}
\end{figure*}

\subsection{Channel Denoising Performance}

The presence of varying fading gains $\bm{H}_z$ and noise with uncertain SNRs $\bm{H}_n\sigma\epsilon$ in wireless channels can severely impair the efficacy of SemCom systems. Accordingly, denoising the noisy signals subsequent to equalization at the receiver emerges as a vital approach to safeguard the desired meaning of the transmitted data. Typically, the channel denoising results of Deep JSCC, JPEG2000+LDPC, VE-LDM, and the proposed EECD methods are demonstrated in Fig. \ref{Fig5} under the conditions of both AWGN and Rayleigh channel at SNR of 10 dB. Herein, the conventional JPEG2000+LDPC approach configures CBR at 1/3 for MNIST and 0.05 for AFHQ/DIV2K datasets for higher performances, whereas the CBR for DL-based methods is set at 1/16 for MNIST, and approximately 0.02 for AFHQ/DIV2K. It is noted that JPEG2000+LDPC suffers from partial bit errors and image blurring at a noise level of 10 dB SNR, resulting in a lower MS-SSIM and a higher LPIPS than DL-based methods.

Advancing further into DL-based methods, SemCom systems constructed on DMs and GANs outperform those based on a CNN-based Deep JSCC approach. As depicted in Fig. \ref{Fig5}, the Deep JSCC method exhibits a slight deficiency in certain image details relative to the latter two methods, leading to marginally inferior semantic metrics. Most crucially, the EECD method, with a subsequence length of $s$=2 used for comparison, demonstrates that the EECD methodology based on VE-LDM distillation virtually matches the performance of the original teacher model at SNR of 10 dB, clearly demonstrating the effectiveness and superiority of the proposed end-to-end human perception metric-based distillation strategy.

\begin{figure*}[!h]
	\centering   
	\subfigure 
	{
        
		\includegraphics[width=0.95\linewidth]{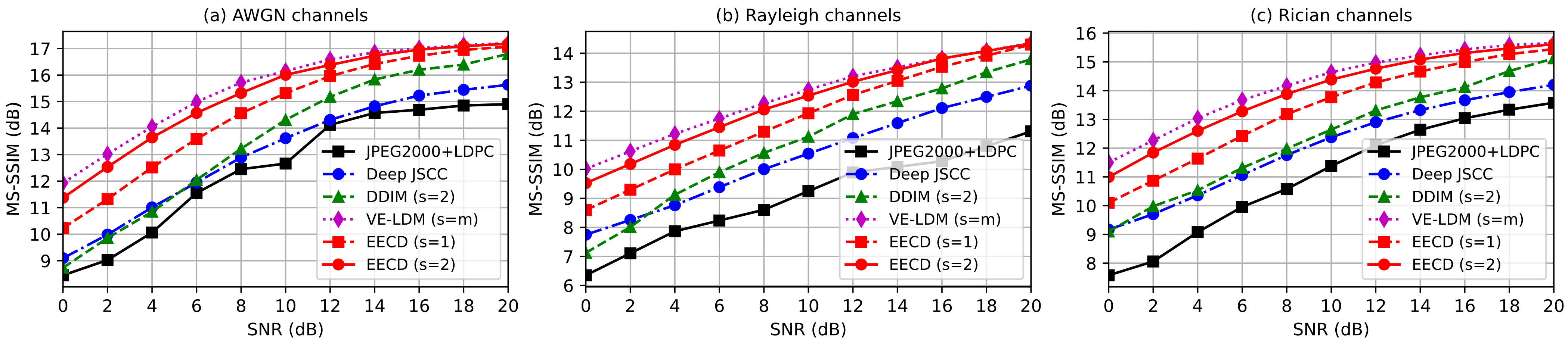}
	}

	\vspace{-0.8em}

	\subfigure
	{
       
		\includegraphics[width=0.95\linewidth]{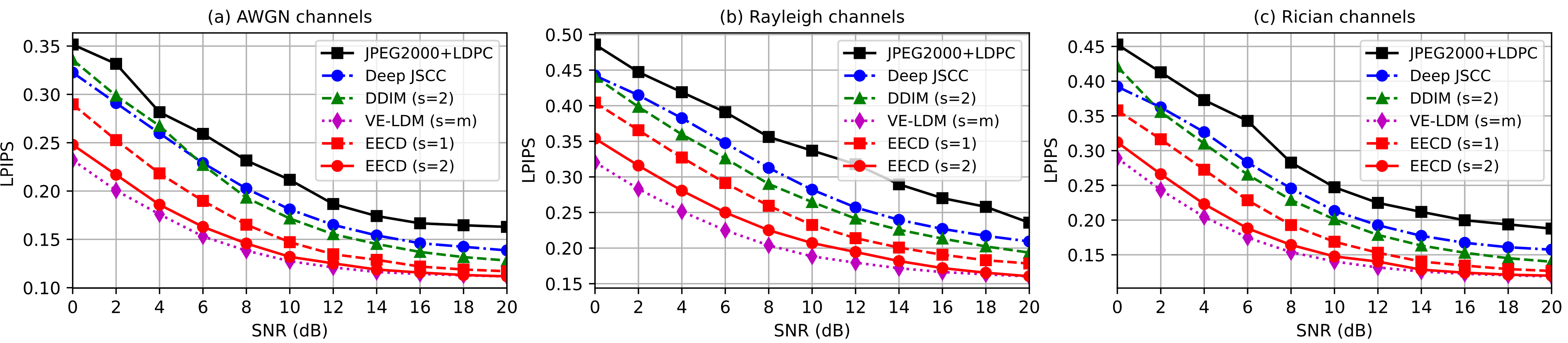}

	}
	\vspace{-0.2cm}
	\caption{Semantic performance metrics of JPEG2000+LDPC, Deep JSCC, DDIM, VE-LDM, and EECD methods under different SNRs and channel states within DIV2K dataset.}
	\label{Fig8}
	\vspace{-0.5cm}
\end{figure*}

Numerical experiments conducted on the AFHQ dataset have provided ample validation for the four distinct methodologies, revealing variations in two pivotal semantic metrics under various channel conditions and noise levels. Specifically, the CBR for JPEG2000+LDPC is set at 0.05, while a unified CBR of 0.02 is employed for the other DL-based methods, and $K$ in Rician channels is 2. Notably, the EECD method employs different subsequence lengths of $s$=2 and $s$=1 to validate its denoising proficiency. As illustrated in Fig. \ref{Fig6}, with the SNR range of 0 dB to 20 dB, a perceptual degradation in quality is observed for all methods in the low-SNR area, with a particularly pronounced decline under Rayleigh and Rician channels, likely induced by fading gains. Evidently, all DL-enabled denoising approaches effectively address the issue of noise sensitivity present in the modulation and demodulation processes of 256-QAM, suppressing the cliff effect found in traditional communication systems while maintaining good semantic accuracy. Conventionally, joint compression and error correction methods exhibit slightly inferior performance compared to DL-based approaches across varying SNRs and channel types, even with higher CBR. Furthermore, the CNN-based Deep JSCC method converges to a different perceptual quality level when compared to methods utilizing DMs and generator as SNR gradually increases. In contrast, VE-LDM and EECD methods converge to the same level of perceptual quality in high-SNR area. Most importantly, the performance of EECD can be further approximated to that of the teacher model, i.e., VE-LDM, with increased resampling length, even in low-SNR area. The experimental results also show that the proposed end-to-end semantic metric-guided consistency training strategy significantly outperforms DDIM in low-SNR conditions with the same number of sampling steps.

Similarly, experiments have been conducted within the DIV2K dataset, the results of which are depicted in Fig. \ref{Fig8}. The CBR and CSI settings for the four methods are consistent with those utilized in the experiments for the AFHQ dataset. Overall, the semantic performance with the DIV2K dataset is slightly inferior to that with the AFHQ dataset. Among these denoising methods, the DM-enabled approaches exhibit the most robustness across different SNR levels, achieving MS-SSIM values of 13-16 dB and LPIPS values of 0.15-0.20 under 10 dB SNR conditions. Specifically, the human-perceptual metrics for the denoising outcomes with the AWGN channel are superior to those with the Rayleigh and Rician channels. In both AWGN and Rician channels, the performance of the denoising methods stabilizes at 20 dB, whereas the performance in the Rayleigh channel continues to fluctuate rapidly as the SNR increases. With regard to different channel denoising approaches, the original channel denoising DM undoubtedly achieves the most favorable performance with $m$ denoising steps, closely followed by the EECD curves with two different subsequence lengths, where the outcomes with $s$=2 are highly proximate to the denoising effects of the original VE-LDM, ensuring the normal transmission of semantic information. Additionally, EECD demonstrates the superiority of distillation and semantic learning compared to DDIM with $s=2$ in low SNR regions, while their performance is similar in high SNR regions.

\begin{figure*}[!h]
	\centering   
	\subfigure 
	{
        
		\includegraphics[width=0.95\linewidth]{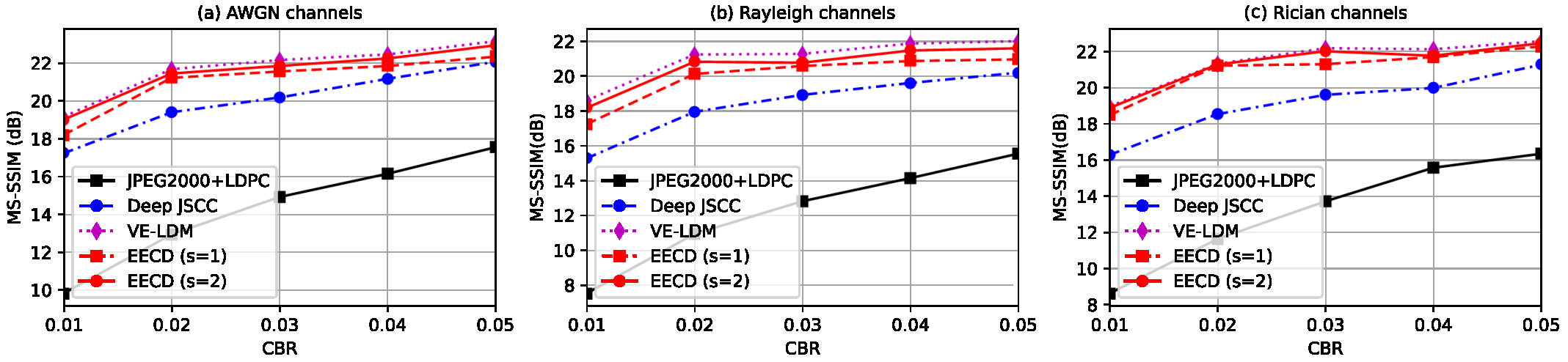}
	}

	\vspace{-0.8em}

	\subfigure
	{
        
		\includegraphics[width=0.95\linewidth]{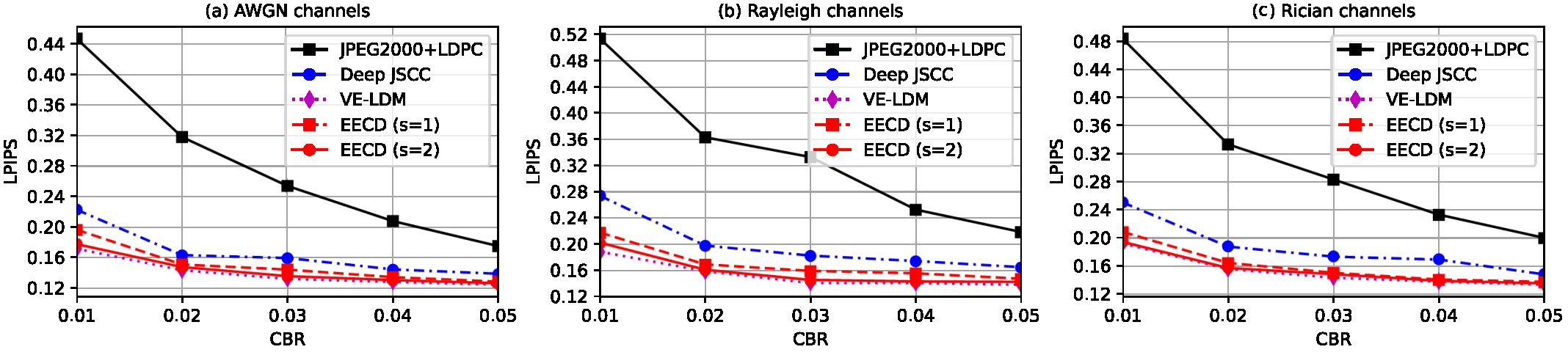}

	}
	\vspace{-0.2cm}
	\caption{Semantic metrics of JPEG2000+LDPC, Deep JSCC, VE-LDM, and EECD methods under different CBRs within AFHQ dataset.}
	\label{Fig10}
	\vspace{-0.5cm}
\end{figure*}

Generally, an exemplary SemCom system is expected to maintain good reconstruction perceptual quality at a lower CBR, ultimately conserving communication bandwidth and reducing the communication burden. For different CBRs, Fig. \ref{Fig10} presents the changes in average perceptual metrics for the four different methods within the AFHQ dataset at CBRs ranging from 0.01 to 0.05. On one hand, the decoding quality of the conventional JPEG2000+LDPC method is heavily influenced by the compression ratio, with different CBRs potentially resulting in a manifold change in perceptual metrics. On the other hand, DL-based methods are less affected by CBR, indicating that DL-based models are robust and excel at extracting data features in the low-CBR area. Moreover, the channel denoising methods constructed based on DMs have attained superior performance under various CBR conditions.

\subsection{Computational Complexity Analysis}

Another paramount requirement for SemCom systems is low-latency communication, encompassing minimal data processing time for encoding, transmission, denoising, and decoding. The introduction of the EECD method enables the distillation of the multi-step denoising process in the latent space of the original DM into a few, or even a one-step sampling process, with only a slight perceptual quality trade-off, thus facilitating real-time SemCom. Specifically, since the VE-LDM and EECD methods both utilize the same robust encoder and generator, only the computational complexity of the denoising process is analyzed. As discussed in \cite{10542391}, the noise prediction computational complexity for the denoising U-Net used by the DM is 
\begin{equation}
	\begin{aligned}
\textrm{Time} \sim \mathcal{O} \left ( \sum^{L}_{l=1} h^2_l w^2_l \cdot C_l \cdot C_{l-1} \cdot K^2_l \right ), 
\end{aligned}
\end{equation}
where $L$ is the number of layers, $h_lw_l$ denotes the feature size and $h_lw_l \propto k$, $C_l$ and $C_{l-1}$ are the number of convolutional kernels in the $l$-th and $l-1$-th layer, and $K_l$ is the edge length of the convolutional kernel in the $l$-th layer. The channel denoising task requires only $m$ NFE based on the noise level, hence the sampling computational complexity is $\textrm{Time} \sim m \times \mathcal{O} \left ( \sum^{L}_{l=1} h^2_l w^2_l \cdot C_l \cdot C_{l-1} \cdot K^2_l \right ) $.  As demonstrated by the time consumptions for different datasets under various CSI conditions in Table \ref{tc_results}, the computing time of VE-LDM may vary dramatically according to the noise level.  However, after the application of EECD, where the denoising steps are fixed at the setting value ($s$=2), the overall time required for the encoding, denoising, and decoding sequence is substantially reduced to mere tens of milliseconds. In comparison with CDDM, if CBR and the number of layers of encoder, decoder and DM are the same, the reduced computational complexity of EECD is $(m-s) \cdot \mathcal{O} \left ( \sum^{L}_{l=1} h^2_l w^2_l \cdot C_l \cdot C_{l-1} \cdot K^2_l \right )$. Inevitably, the proposed method increases the computational complexity of  and the number of layers of encoder, decoder and DM are the same, the reduced computational complexity of EECD is $s \cdot \mathcal{O} \left ( \sum^{L}_{l=1} h^2_l w^2_l \cdot C_l \cdot C_{l-1} \cdot K^2_l \right )$ compared to Deep JSCC.

\begin{figure}[!h]
	\vspace{-0.1cm}
	\centerline{\includegraphics[width=0.48\textwidth]{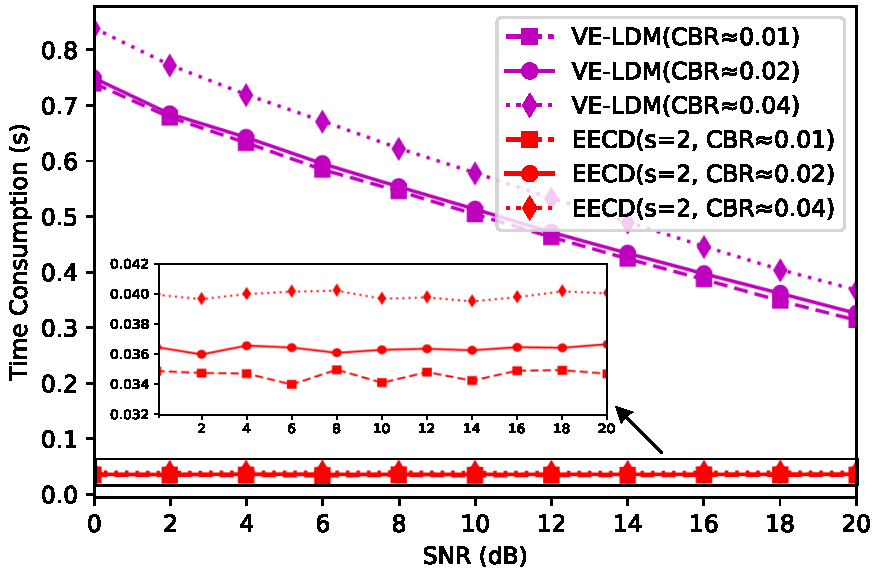}}
	\caption{Time consumptions of single image's encoding, denoising, and decoding process when utilizing VE-LDM and EECD methods under different CBRs and CSIs.  }
	\label{Fig12}
	\vspace{-0.2cm}
\end{figure}

\begin{table}[!h]\centering \small
	\caption{Time consumptions of VE-LDM and EECD methods under different CSIs and datasets (VE-LDM/EECD (miliseconds))}
	\label{tc_results}
	\renewcommand{\arraystretch}{1.1}
	\setlength\tabcolsep{0.7em}
	\begin{tabular}{ccccc}
	\toprule
	\toprule
	CSI  & SNR (dB) & MNIST & AFHQ & DIV2K    \\
	\midrule
	\multirow{3}{*}{AWGN}      & 0   &  762.3/32.7     &   749.6/36.4   &   832.6/43.0  \\
    & 10   &  543.5/33.0     &  513.6/36.3    &  591.3/43.2  \\
	& 20   &  361.7/32.9     &  326.4/36.6    &  396.7/43.1  \\
    \midrule
	\multirow{3}{*}{Rayleigh}      & 0   &  762.7/32.5     &  750.1/36.5    &  833.1/43.1   \\
    & 10   &  543.7/33.2     &  514.1/36.3    &  590.9/43.0  \\
	& 20   &  362.3/33.0     &  327.8/36.8    &  397.1/43.3  \\
	\midrule
	\multirow{3}{*}{Rician}      & 0   &  761.9/32.5     &  749.7/36.4    &   832.8/43.2\\
    & 10   &  542.9/32.9     &  513.7/36.4    &  591.4/43.3  \\
	& 20   &  362.1/32.4     &  327.0/36.5    &  396.9/43.1  \\	
	\bottomrule
	\bottomrule
	\end{tabular}
	\vspace{-0.1cm}
	\end{table}

Within the AFHQ dataset, the time consumption variability of VE-LDM and EECD models trained at different CBRs across an SNR range from  0 dB to 20 dB is illustrated in Fig. \ref{Fig12}. In conjunction with the data presented in Table \ref{tc_results} and Fig. \ref{Fig12}, it is evident that both the CBR and the image resolution can influence the channel denoising processing time. Consequently, VE-LDM may not meet the real-time denoising requirements in scenarios with low-SNR, high-CBR, or high-resolution. However, the predominant factor affecting the VE-LDM during the denoising process is the noise level. Consequently, in contrast to the VE-LDM's more substantial and variable time complexity during denoising, the proposed EECD method consistently maintains the time required for the denoising task within the scale of tens of milliseconds. Additionally, in line with the numerical results previously discussed, EECD does not significantly degrade semantic quality across various CSI scenarios.

In summary, the proposed SemCom system achieves a good balance between latency and robustness. The VAE (6 convolutional layers) and GAN models (7 deconvolutional layers), with average encoding and decoding times of 6.21 ms and 8.33 ms, respectively, for a single AFHQ image, introduce minimal decoding computational burden at the resource-rich transmitter. When significant semantic ambiguity is detected, the activation of one-shot learning with an average duration of approximately 450 ms ensures a rapid improvement in the quality of \textit{out-of-domain} images. Fortunately, these strategies can be further enhanced by integrating advanced edge-cloud collaborative methods \cite{10508191} and optimized encoding/decoding mechanisms \cite{10689268}.


\section{Conclusion} 
\label{sec:Conclusion}
This paper introduces a wireless semantic communication (SemCom) system tailored to navigate the challenges of semantic ambiguities and channel noises. The proposed SemCom system's proficiency in feature extraction diminishes the adverse effects of outliers in source data on deep learning-based communication systems and exhibits an impressive aptitude for rapid adaptation to data with unknown distribution, thereby augmenting the human-perceptual quality of decoded data. In the realm of data transmission, the advanced end-to-end consistency distillation (EECD) strategy facilitates real-time channel denoising across various pre-estimated channel state information (CSI) scenarios, achieving this with minimal perceptual quality degradation when contrasted with the existing channel denoising diffusion model techniques. Nonetheless, the real-time SemCom system based on diffusion models with unknown CSI, images with ultra-high resolution (2K/4K/6K), and large network environments still warrants further investigation. Additionally, the integration of diffusion models into next-generation communication paradigms, specifically goal/task-oriented SemCom systems, poses an intriguing and significant topic for future exploration.


\appendix

\subsection{VUB Transformation for VAE-WGAN} \label{transVAEWGAN}

Proof of Eq. \eqref{LJSCC2}:
\begin{equation} 
	\begin{aligned}
		&\mathbb{E}_{q_{\bm{\phi}}(\bm{z}|\bm{x})}\left [ - \log p_{\bm{\psi}}(\bm{x}|\bm{z})\right ] + \mathbb{E}_{q_{\bm{\phi}}(\bm{z}|\bm{x})}\left [  \log q_{\bm{\phi}}(\bm{z}|\bm{x})\right ]\\
		&= \mathbb{E}_{q_{\bm{\phi}}(\bm{z}|\bm{x})}\left [ - \log p_{\bm{\psi}}(\bm{x}|\bm{z})\right ] + \underbrace{\mathbb{E}_{q_{\bm{\phi}}(\bm{z}|\bm{x})}\left [  \log \frac{q_{\bm{\phi}}(\bm{z}|\bm{x})}{p_{\bm{\psi}}(\bm{z})} \right ]}_{\mathbb{E}_q\left [  \mathcal{D}_{KL}(q_{\bm{\phi}}(\bm{z}|\bm{x})\parallel p_{\bm{\psi}}(\bm{z}) )\right ]} \\
		& \quad + \mathbb{E}_{q_{\bm{\phi}}(\bm{z}|\bm{x})}\left [ \log p_{\bm{\psi}}(\bm{z})\right ]\\
		&\geq \mathbb{E}_{q_{\bm{\phi}}(\bm{z}|\bm{x})}\left [ \log \frac{q_{\bm{\phi}}(\bm{z}|\bm{x})}{p_{\bm{\psi}}(\bm{z})}  - \log p_{\bm{\psi}}(\bm{x}|\bm{z})  \right ]\\
		& = \int q_{\bm{\phi}}(\bm{z}|\bm{x}) \log \frac{q_{\bm{\phi}}(\bm{z}|\bm{x})}{p_{\bm{\psi}}(\bm{x}|\bm{z})p_{\bm{\psi}}(\bm{z})} d \bm{z} \\
		& = \int q_{\bm{\phi}}(\bm{z}|\bm{x}) \left [\log p_{\bm{\psi}}(\bm{x})  +\log \frac{q_{\bm{\phi}}(\bm{z}|\bm{x})}{p_{\bm{\psi}}(\bm{z}, \bm{x})} \right ] d \bm{z} - \mathbb{E}_q\left [  \log p_{\bm{\psi}}(\bm{x}) \right ]\\
		& = \int q_{\bm{\phi}}(\bm{z}|\bm{x}) \left [\log \frac{q_{\bm{\phi}}(\bm{z}|\bm{x})}{p_{\bm{\psi}}(\bm{z}| \bm{x})} \right ] d \bm{z}+ \mathbb{E}_q\left [-  \log p_{\bm{\psi}}(\bm{x}) \right ]\\
		& = \mathbb{E}_{q(\bm{x})}\left [ \mathcal{D}_{KL}\left (  q_{\bm{\phi}}(\bm{z}|\bm{x})\parallel p_{\bm{\psi}}(\bm{z}| \bm{x}) \right ) \right ] + \mathbb{E}_{q_{\bm{\psi}}(\bm{z})}\left [-  \log p_{\bm{\psi}}(\bm{x}) \right ].
	\end{aligned}
\end{equation}

\subsection{Training Process of VAE-WGAN-GP} \label{trainingVAEWGANGP}

The training process of deep CNN based VAE-WGAN with gradient penalty \cite{gulrajani2017improved} is illustrated in Algorithm \ref{trainingVAEWGAN}, where $\alpha_{\bm{\phi}}$ and $\alpha_{\bm{\psi}}$ are the loss balance hyperparameters.

\begin{algorithm}[!h]  	\label{trainingVAEWGAN}
	\small
	\caption{Training algorithm of VAE-WGAN-GP}
		\LinesNumbered
		\KwIn{ Dataset $q(\bm{x})$, learning rate $\eta$, gradient penalty coefficient $\lambda$, loss balance hyperparameters $\alpha_{\bm{\phi}}$ and $\alpha_{\bm{\psi}}$, the number of iterations $n_{critic}$ of discriminator per generator iteration, initial encoder parameter $\bm{\phi}$, generator parameter $\bm{\psi}$, discriminator parameter $\bm{\gamma}$}
		\KwOut{The trained $E_{\bm{\phi}}(.)$, $G_{\bm{\psi}}(.)$, and $D_{\bm{\gamma}}(.)$}
		\Repeat{Converged}{
        \For{$i=0$, $\cdots$, $n_{critic}$}{
            Sample $\bm{x} \sim q(\bm{x})$, $\bm{z} \sim q_{\bm{\psi}}(\bm{z})$, and $\epsilon_1,\epsilon_2 \sim U[0,1]$\;
            Compute $\hat{\bm{x}}_1 \leftarrow \epsilon_1\bm{x} + (1-\epsilon_1)G_{\bm{\psi}}(\bm{z})$\;
            Compute $\hat{\bm{x}}_2 \leftarrow \epsilon_2\bm{x} + (1-\epsilon_2)G_{\bm{\psi}}(E_{\bm{\phi}}(\bm{x}))$\;
			Update $\bm{\gamma}$ by $\bm{\gamma} \leftarrow \bm{\gamma} - \eta \nabla_{\bm{\gamma}} \Big [  \mathbb{E}_q\big (- 2 D_{\bm{\gamma}(\bm{x})} + D_{\bm{\gamma}}(G_{\bm{\psi}}(\bm{z}))+ D_{\bm{\gamma}}(G_{\bm{\psi}}(E_{\bm{\phi}}(\bm{x}))) + \lambda (\left \| \nabla_{\hat{\bm{x}}_1} D_{\bm{\gamma}}(\hat{\bm{x}}_1)  \right \|_2 -1)^2 + \lambda (\left \| \nabla_{\hat{\bm{x}}_2}D_{\bm{\gamma}}(\hat{\bm{x}}_2)  \right \|_2  -1 )^2  \big ) \Big ]$ \;
			
		}
		Sample $\bm{x} \sim q(\bm{x})$ and $\bm{z} \sim q_{\bm{\psi}}(\bm{z})$\;
		Update $\bm{\phi}$ by $\bm{\phi} \leftarrow \bm{\phi} - \eta \nabla_{\bm{\phi}}\Big [ \mathbb{E}_q \Big (  \alpha_{\bm{\phi}}\mathcal{D}_{KL}\big ( E_{\bm{\phi}}(\bm{x})\sim \mathcal{N}(\bm{\mu}, \bm{\sigma}^2) \parallel \bm{z} \sim \mathcal{N}(\bm{0}, \bm{I}) \big ) + (1- \alpha_{\bm{\phi}})\mathcal{D}_{KL}\left (G_{\bm{\psi}}(E_{\bm{\phi}}(\bm{x})) \parallel \bm{x} \right ) \Big ) \Big ]$\;
		Update $\bm{\psi}$ by $\bm{\psi} \leftarrow \bm{\psi} -\eta \nabla_{\bm{\phi}}\Big [  \mathbb{E}_q\Big ( \alpha_{\bm{\phi}} \mathcal{D}_{KL}\left ( G_{\bm{\psi}}(E_{\bm{\phi}} (\bm{x}))\parallel \bm{x} \right ) + (1-\alpha_{\bm{\psi}})\left ( - D_{\bm{\gamma}}(G_{\bm{\psi}}(\bm{z})) - D_{\bm{\gamma}}(G_{\bm{\psi}}(E_{\bm{\phi}}(\bm{x})))\right )  \Big ) \Big ]$\;
		}
		\textbf{Return} Trained VAE-WGAN-GP Model
\end{algorithm}

\subsection{Proof of Robust Encoder's VUB} \label{proofVUB}

According to conditional Markov random field model \cite{sutton2012introduction}, the joint distribution of data $\bm{x}$ and data with semantic error $\bm{x}+\bm{\delta}$ is
\begin{equation} 
	\begin{aligned}
        &p_{\bm{\psi}}(\bm{x}, \bm{x}+\bm{\delta}) \propto  \\
		&\int  p_{\bm{\psi}}(\bm{x}|\bm{z})p_{\bm{\psi}}(\bm{x}+\bm{\delta}|\bm{z}'')e^{-\frac{\beta}{2}\bm{d}(\bm{z},\bm{z}'')}p(\bm{z})p(\bm{z}'')d\bm{z}d\bm{z}'',
	\end{aligned}
\end{equation}
where $\beta$ denotes the nonnegative coupling parameter. Consequently, considering the joint distribution $q(\bm{z}, \bm{z}'')$, the evidence lower bound has the following form
\begin{equation} \label{preq}
	\begin{aligned}
&\mathbb{E}_q\left [  \log  p_{\bm{\psi}}(\bm{x}, \bm{x}+\bm{\delta}) \right ] \geq \mathbb{E}_{q(\bm{z})}\left [\log p_{\bm{\psi}}(\bm{x}|\bm{z})\right ] + \mathbb{E}_{q(\bm{z})}\left [\log p_{\bm{\psi}}(\bm{z})\right ]\\
& +\mathbb{E}_{q(\bm{z}'')}\left [\log p_{\bm{\psi}}(\bm{x}+\bm{\delta}|\bm{z}'')\right ] +\mathbb{E}_{q(\bm{z}'')}\left [\log p_{\bm{\psi}}(\bm{z}'')\right ] \\
&- \frac{\beta}{2}\mathbb{E}_{q(\bm{z}, \bm{z}'')}\bm{d}(\bm{z}, \bm{z}'')  +\mathbb{E}_{q(\bm{z}, \bm{z}'')} \left [ \log q(\bm{z}, \bm{z}'')\right ].
	\end{aligned}
\end{equation}
To decode clean data without changing the decoder parameters $\bm{\psi}$, by simply integrating out $\bm{x}+\bm{\delta}$ in Eq. \eqref{preq} and according to Jensen's inequality, term $\mathbb{E}_q\left [ - \log p_{\bm{\psi}}(\bm{x}) \right ]$ can be transformed into Eq. \eqref{vlbr}.

\bibliographystyle{IEEEtran}
\bibliography{IEEEabrv,bibi}

\begin{thebibliography}{10}
\providecommand{\url}[1]{#1}
\csname url@samestyle\endcsname
\providecommand{\newblock}{\relax}
\providecommand{\bibinfo}[2]{#2}
\providecommand{\BIBentrySTDinterwordspacing}{\spaceskip=0pt\relax}
\providecommand{\BIBentryALTinterwordstretchfactor}{4}
\providecommand{\BIBentryALTinterwordspacing}{\spaceskip=\fontdimen2\font plus
\BIBentryALTinterwordstretchfactor\fontdimen3\font minus
  \fontdimen4\font\relax}
\providecommand{\BIBforeignlanguage}[2]{{%
\expandafter\ifx\csname l@#1\endcsname\relax
\typeout{** WARNING: IEEEtran.bst: No hyphenation pattern has been}%
\typeout{** loaded for the language `#1'. Using the pattern for}%
\typeout{** the default language instead.}%
\else
\language=\csname l@#1\endcsname
\fi
#2}}
\providecommand{\BIBdecl}{\relax}
\BIBdecl

\bibitem{siriwardhana2021survey}
Y.~Siriwardhana \emph{et~al.}, ``A survey on mobile augmented reality with 5g
  mobile edge computing: Architectures, applications, and technical aspects,''
  \emph{IEEE Communications Surveys \& Tutorials}, vol.~23, no.~2, pp.
  1160--1192, 2021.

\bibitem{huang2023virtual}
X.~Huang, J.~Riddell, and R.~Xiao, ``Virtual reality telepresence: 360-degree
  video streaming with edge-compute assisted static foveated compression,''
  \emph{IEEE Transactions on Visualization and Computer Graphics}, 2023.

\bibitem{abrahamsen2021communication}
F.~E. Abrahamsen, Y.~Ai, and M.~Cheffena, ``Communication technologies for
  smart grid: A comprehensive survey,'' \emph{Sensors}, vol.~21, no.~23, p.
  8087, 2021.

\bibitem{wu2022unmanned}
W.~Wu \emph{et~al.}, ``Unmanned aerial vehicle swarm-enabled edge computing:
  Potentials, promising technologies, and challenges,'' \emph{IEEE Wireless
  Communications}, vol.~29, no.~4, pp. 78--85, 2022.

\bibitem{chowdhury20206g}
M.~Z. Chowdhury \emph{et~al.}, ``6g wireless communication systems:
  Applications, requirements, technologies, challenges, and research
  directions,'' \emph{IEEE Open Journal of the Communications Society}, vol.~1,
  pp. 957--975, 2020.

\bibitem{8723589}
E.~Bourtsoulatze, D.~Burth~Kurka, and D.~Gündüz, ``Deep joint source-channel
  coding for wireless image transmission,'' \emph{IEEE Transactions on
  Cognitive Communications and Networking}, vol.~5, no.~3, pp. 567--579, 2019.

\bibitem{wallace1992jpeg}
G.~K. Wallace, ``The {JPEG} still picture compression standard,'' \emph{IEEE
  transactions on consumer electronics}, vol.~38, no.~1, pp. xviii--xxxiv,
  1992.

\bibitem{christopoulos2000jpeg2000}
C.~Christopoulos, A.~Skodras, and T.~Ebrahimi, ``The {JPEG2000} still image
  coding system: an overview,'' \emph{IEEE transactions on consumer
  electronics}, vol.~46, no.~4, pp. 1103--1127, 2000.

\bibitem{fan2018wide}
Y.~Fan, J.~Yu, and T.~S. Huang, ``Wide-activated deep residual networks based
  restoration for {BPG}-compressed images,'' in \emph{Proceedings of the IEEE
  conference on computer vision and pattern recognition Workshops}, 2018, pp.
  2621--2624.

\bibitem{yang2022semantic}
W.~Yang \emph{et~al.}, ``Semantic communications for future internet:
  Fundamentals, applications, and challenges,'' \emph{IEEE Communications
  Surveys \& Tutorials}, vol.~25, no.~1, pp. 213--250, 2022.

\bibitem{luo2022semantic}
X.~Luo, H.-H. Chen, and Q.~Guo, ``Semantic communications: Overview, open
  issues, and future research directions,'' \emph{IEEE Wireless
  Communications}, vol.~29, no.~1, pp. 210--219, 2022.

\bibitem{xie2021deep}
H.~Xie \emph{et~al.}, ``Deep learning enabled semantic communication systems,''
  \emph{IEEE Transactions on Signal Processing}, vol.~69, pp. 2663--2675, 2021.

\bibitem{dai2022nonlinear}
J.~Dai \emph{et~al.}, ``Nonlinear transform source-channel coding for semantic
  communications,'' \emph{IEEE Journal on Selected Areas in Communications},
  vol.~40, no.~8, pp. 2300--2316, 2022.

\bibitem{he2016deep}
K.~He \emph{et~al.}, ``Deep residual learning for image recognition,'' in
  \emph{Proceedings of the IEEE conference on computer vision and pattern
  recognition}, 2016, pp. 770--778.

\bibitem{xu2021wireless}
J.~Xu \emph{et~al.}, ``Wireless image transmission using deep source channel
  coding with attention modules,'' \emph{IEEE Transactions on Circuits and
  Systems for Video Technology}, vol.~32, no.~4, pp. 2315--2328, 2021.

\bibitem{farsad2018deep}
N.~Farsad, M.~Rao, and A.~Goldsmith, ``Deep learning for joint source-channel
  coding of text,'' in \emph{2018 IEEE international conference on acoustics,
  speech and signal processing (ICASSP)}.\hskip 1em plus 0.5em minus
  0.4em\relax IEEE, 2018, pp. 2326--2330.

\bibitem{9953099}
H.~Zhang \emph{et~al.}, ``Deep learning-enabled semantic communication systems
  with task-unaware transmitter and dynamic data,'' \emph{IEEE Journal on
  Selected Areas in Communications}, vol.~41, no.~1, pp. 170--185, 2023.

\bibitem{huang2022toward}
D.~Huang \emph{et~al.}, ``Toward semantic communications: Deep learning-based
  image semantic coding,'' \emph{IEEE Journal on Selected Areas in
  Communications}, vol.~41, no.~1, pp. 55--71, 2022.

\bibitem{weng2021semantic}
Z.~Weng and Z.~Qin, ``Semantic communication systems for speech transmission,''
  \emph{IEEE Journal on Selected Areas in Communications}, vol.~39, no.~8, pp.
  2434--2444, 2021.

\bibitem{xie2022task}
H.~Xie \emph{et~al.}, ``Task-oriented multi-user semantic communications,''
  \emph{IEEE Journal on Selected Areas in Communications}, vol.~40, no.~9, pp.
  2584--2597, 2022.

\bibitem{ho2020denoising}
J.~Ho, A.~Jain, and P.~Abbeel, ``Denoising diffusion probabilistic models,''
  \emph{Advances in neural information processing systems}, vol.~33, pp.
  6840--6851, 2020.

\bibitem{song2020denoising}
J.~Song, C.~Meng, and S.~Ermon, ``Denoising diffusion implicit models,''
  \emph{arXiv preprint arXiv:2010.02502}, 2020.

\bibitem{du2023beyond}
H.~Du \emph{et~al.}, ``Enhancing deep reinforcement learning: A tutorial on
  generative diffusion models in network optimization,'' \emph{IEEE
  Communications Surveys \& Tutorials}, 2024.

\bibitem{qiao2024latency}
L.~Qiao \emph{et~al.}, ``Latency-aware generative semantic communications with
  pre-trained diffusion models,'' \emph{IEEE Wireless Communications Letters},
  vol.~13, no.~10, pp. 2652--2656, 2024.

\bibitem{du2023ai}
H.~Du \emph{et~al.}, ``Ai-generated incentive mechanism and full-duplex
  semantic communications for information sharing,'' \emph{IEEE Journal on
  Selected Areas in Communications}, vol.~41, no.~9, pp. 2981--2997, 2023.

\bibitem{chen2024commin}
J.~Chen \emph{et~al.}, ``Commin: Semantic image communications as an inverse
  problem with inn-guided diffusion models,'' in \emph{ICASSP 2024-2024 IEEE
  International Conference on Acoustics, Speech and Signal Processing
  (ICASSP)}.\hskip 1em plus 0.5em minus 0.4em\relax IEEE, 2024, pp. 6675--6679.

\bibitem{grassucci2023generative}
E.~Grassucci, S.~Barbarossa, and D.~Comminiello, ``Generative semantic
  communication: Diffusion models beyond bit recovery,'' \emph{arXiv preprint
  arXiv:2306.04321}, 2023.

\bibitem{yilmaz2023high}
S.~F. Yilmaz \emph{et~al.}, ``High perceptual quality wireless image delivery
  with denoising diffusion models,'' in \emph{IEEE INFOCOM 2024-IEEE Conference
  on Computer Communications Workshops (INFOCOM WKSHPS)}.\hskip 1em plus 0.5em
  minus 0.4em\relax IEEE, 2024, pp. 1--5.

\bibitem{yang2024sg2sc}
M.~Yang \emph{et~al.}, ``{SG2SC}: A generative semantic communication framework
  for scene understanding-oriented image transmission,'' in \emph{ICASSP
  2024-2024 IEEE International Conference on Acoustics, Speech and Signal
  Processing (ICASSP)}.\hskip 1em plus 0.5em minus 0.4em\relax IEEE, 2024, pp.
  13\,486--13\,490.

\bibitem{choukroun2022denoising}
Y.~Choukroun and L.~Wolf, ``Denoising diffusion error correction codes,''
  \emph{arXiv preprint arXiv:2209.13533}, 2022.

\bibitem{zilberstein2024joint}
N.~Zilberstein, A.~Swami, and S.~Segarra, ``Joint channel estimation and data
  detection in massive mimo systems based on diffusion models,'' in
  \emph{ICASSP 2024-2024 IEEE International Conference on Acoustics, Speech and
  Signal Processing (ICASSP)}.\hskip 1em plus 0.5em minus 0.4em\relax IEEE,
  2024, pp. 13\,291--13\,295.

\bibitem{jiang2024large}
F.~Jiang \emph{et~al.}, ``Large generative model assisted {3D} semantic
  communication,'' \emph{arXiv preprint arXiv:2403.05783}, 2024.

\bibitem{jiang2024diffsc}
Z.~Jiang \emph{et~al.}, ``{DIFFSC}: Semantic communication framework with
  enhanced denoising through diffusion probabilistic models,'' in \emph{ICASSP
  2024-2024 IEEE International Conference on Acoustics, Speech and Signal
  Processing (ICASSP)}.\hskip 1em plus 0.5em minus 0.4em\relax IEEE, 2024, pp.
  13\,071--13\,075.

\bibitem{grassucci2024diffusion}
E.~Grassucci \emph{et~al.}, ``Diffusion models for audio semantic
  communication,'' in \emph{ICASSP 2024-2024 IEEE International Conference on
  Acoustics, Speech and Signal Processing (ICASSP)}.\hskip 1em plus 0.5em minus
  0.4em\relax IEEE, 2024, pp. 13\,136--13\,140.

\bibitem{du2023exploring}
H.~Du \emph{et~al.}, ``Exploring collaborative distributed diffusion-based
  ai-generated content (aigc) in wireless networks,'' \emph{IEEE Network},
  vol.~38, no.~3, pp. 178--186, 2023.

\bibitem{10480348}
T.~Wu \emph{et~al.}, ``Cddm: Channel denoising diffusion models for wireless
  semantic communications,'' \emph{IEEE Transactions on Wireless
  Communications}, vol.~23, no.~9, pp. 11\,168--11\,183, 2024.

\bibitem{10104549}
M.~Kim, R.~Fritschek, and R.~F. Schaefer, ``Learning end-to-end channel coding
  with diffusion models,'' in \emph{WSA \& SCC 2023; 26th International ITG
  Workshop on Smart Antennas and 13th Conference on Systems, Communications,
  and Coding}, 2023, pp. 1--13.

\bibitem{10542391}
J.~Pei \emph{et~al.}, ``Detection and imputation based two-stage denoising
  diffusion power system measurement recovery under cyber-physical
  uncertainties,'' \emph{IEEE Transactions on Smart Grid}, pp. 1--1, 2024.

\bibitem{zhang2022unified}
G.~Zhang \emph{et~al.}, ``A unified multi-task semantic communication system
  with domain adaptation,'' in \emph{GLOBECOM 2022-2022 IEEE Global
  Communications Conference}.\hskip 1em plus 0.5em minus 0.4em\relax IEEE,
  2022, pp. 3971--3976.

\bibitem{rombach2022high}
R.~Rombach \emph{et~al.}, ``High-resolution image synthesis with latent
  diffusion models,'' in \emph{Proceedings of the IEEE/CVF conference on
  computer vision and pattern recognition}, 2022, pp. 10\,684--10\,695.

\bibitem{adler2018banach}
J.~Adler and S.~Lunz, ``Banach {W}asserstein {GAN},'' \emph{Advances in neural
  information processing systems}, vol.~31, 2018.

\bibitem{song2023consistency}
Y.~Song \emph{et~al.}, ``Consistency models,'' \emph{arXiv preprint
  arXiv:2303.01469}, 2023.

\bibitem{chen2005reduced}
J.~Chen \emph{et~al.}, ``Reduced-complexity decoding of {LDPC} codes,''
  \emph{IEEE transactions on communications}, vol.~53, no.~8, pp. 1288--1299,
  2005.

\bibitem{adesina2022adversarial}
D.~Adesina \emph{et~al.}, ``Adversarial machine learning in wireless
  communications using {RF} data: A review,'' \emph{IEEE Communications Surveys
  \& Tutorials}, vol.~25, no.~1, pp. 77--100, 2022.

\bibitem{liu2020deep}
Y.~Liu \emph{et~al.}, ``Deep anomaly detection for time-series data in
  industrial iot: A communication-efficient on-device federated learning
  approach,'' \emph{IEEE Internet of Things Journal}, vol.~8, no.~8, pp.
  6348--6358, 2020.

\bibitem{10416926}
G.~Zheng \emph{et~al.}, ``Mobility-aware split-federated with transfer learning
  for vehicular semantic communication networks,'' \emph{IEEE Internet of
  Things Journal}, pp. 1--1, 2024.

\bibitem{nozza2016deep}
D.~Nozza, E.~Fersini, and E.~Messina, ``Deep learning and ensemble methods for
  domain adaptation,'' in \emph{2016 IEEE 28th International conference on
  tools with artificial intelligence (ICTAI)}.\hskip 1em plus 0.5em minus
  0.4em\relax IEEE, 2016, pp. 184--189.

\bibitem{khan2019robust}
F.~N. Khan and A.~P.~T. Lau, ``Robust and efficient data transmission over
  noisy communication channels using stacked and denoising autoencoders,''
  \emph{China Communications}, vol.~16, no.~8, pp. 72--82, 2019.

\bibitem{ye2020deep}
H.~Ye \emph{et~al.}, ``Deep learning-based end-to-end wireless communication
  systems with conditional gans as unknown channels,'' \emph{IEEE Transactions
  on Wireless Communications}, vol.~19, no.~5, pp. 3133--3143, 2020.

\bibitem{vahdat2021score}
A.~Vahdat, K.~Kreis, and J.~Kautz, ``Score-based generative modeling in latent
  space,'' \emph{Advances in neural information processing systems}, vol.~34,
  pp. 11\,287--11\,302, 2021.

\bibitem{abdal2019image2stylegan}
R.~Abdal, Y.~Qin, and P.~Wonka, ``Image2stylegan: How to embed images into the
  stylegan latent space?'' in \emph{Proceedings of the IEEE/CVF international
  conference on computer vision}, 2019, pp. 4432--4441.

\bibitem{wang2022diffusion}
Z.~Wang \emph{et~al.}, ``Diffusion-{GAN}: Training {GAN}s with diffusion,''
  \emph{arXiv preprint arXiv:2206.02262}, 2022.

\bibitem{xia2022gan}
W.~Xia \emph{et~al.}, ``{GAN} inversion: A survey,'' \emph{IEEE transactions on
  pattern analysis and machine intelligence}, vol.~45, no.~3, pp. 3121--3138,
  2022.

\bibitem{larsen2016autoencoding}
A.~B.~L. Larsen \emph{et~al.}, ``Autoencoding beyond pixels using a learned
  similarity metric,'' in \emph{International conference on machine
  learning}.\hskip 1em plus 0.5em minus 0.4em\relax PMLR, 2016, pp. 1558--1566.

\bibitem{lanfredi2023quantifying}
R.~B. Lanfredi, J.~D. Schroeder, and T.~Tasdizen, ``Quantifying the
  preferential direction of the model gradient in adversarial training with
  projected gradient descent,'' \emph{Pattern Recognition}, vol. 139, p.
  109430, 2023.

\bibitem{cemgil2019adversarially}
T.~Cemgil \emph{et~al.}, ``Adversarially robust representations with smooth
  encoders,'' in \emph{International Conference on Learning Representations},
  2020, pp. 1--18.

\bibitem{10185192}
P.~R. Gautam, L.~Zhang, and P.~Fan, ``Hybrid {MMSE} precoding for millimeter
  wave {MU-MISO} via trace maximization,'' \emph{IEEE Transactions on Wireless
  Communications}, vol.~23, no.~3, pp. 1999--2010, 2024.

\bibitem{karras2022elucidating}
T.~Karras \emph{et~al.}, ``Elucidating the design space of diffusion-based
  generative models,'' \emph{Advances in Neural Information Processing
  Systems}, vol.~35, pp. 26\,565--26\,577, 2022.

\bibitem{zhou2023fast}
Z.~Zhou \emph{et~al.}, ``Fast {ODE}-based sampling for diffusion models in
  around 5 steps,'' \emph{arXiv preprint arXiv:2312.00094}, 2023.

\bibitem{zhang2018unreasonable}
R.~Zhang \emph{et~al.}, ``The unreasonable effectiveness of deep features as a
  perceptual metric,'' in \emph{Proceedings of the IEEE conference on computer
  vision and pattern recognition}, 2018, pp. 586--595.

\bibitem{choi2020stargan}
Y.~Choi \emph{et~al.}, ``Stargan v2: Diverse image synthesis for multiple
  domains,'' in \emph{Proceedings of the IEEE/CVF conference on computer vision
  and pattern recognition}, 2020, pp. 8188--8197.

\bibitem{Timofte_2018_CVPR_Workshops}
R.~Timofte \emph{et~al.}, ``{NTIRE} 2018 challenge on single image
  super-resolution: Methods and results,'' in \emph{The IEEE Conference on
  Computer Vision and Pattern Recognition (CVPR) Workshops}, June 2018.

\bibitem{10508191}
Y.~Wang \emph{et~al.}, ``End-edge-cloud collaborative computing for deep
  learning: A comprehensive survey,'' \emph{IEEE Communications Surveys \&
  Tutorials}, vol.~26, no.~4, pp. 2647--2683, 2024.

\bibitem{10689268}
C.~Cai, X.~Yuan, and Y.-J. Angela~Zhang, ``Multi-device task-oriented
  communication via maximal coding rate reduction,'' \emph{IEEE Transactions on
  Wireless Communications}, vol.~23, no.~12, pp. 18\,096--18\,110, 2024.

\bibitem{gulrajani2017improved}
I.~Gulrajani \emph{et~al.}, ``Improved training of {W}asserstein {GAN}s,''
  \emph{Advances in neural information processing systems}, vol.~30, 2017.

\bibitem{sutton2012introduction}
C.~Sutton, A.~McCallum \emph{et~al.}, ``An introduction to conditional random
  fields,'' \emph{Foundations and Trends{\textregistered} in Machine Learning},
  vol.~4, no.~4, pp. 267--373, 2012.

\end{thebibliography}

\begin{IEEEbiography}[{\includegraphics[width=1in,height=1.25in,clip,keepaspectratio]{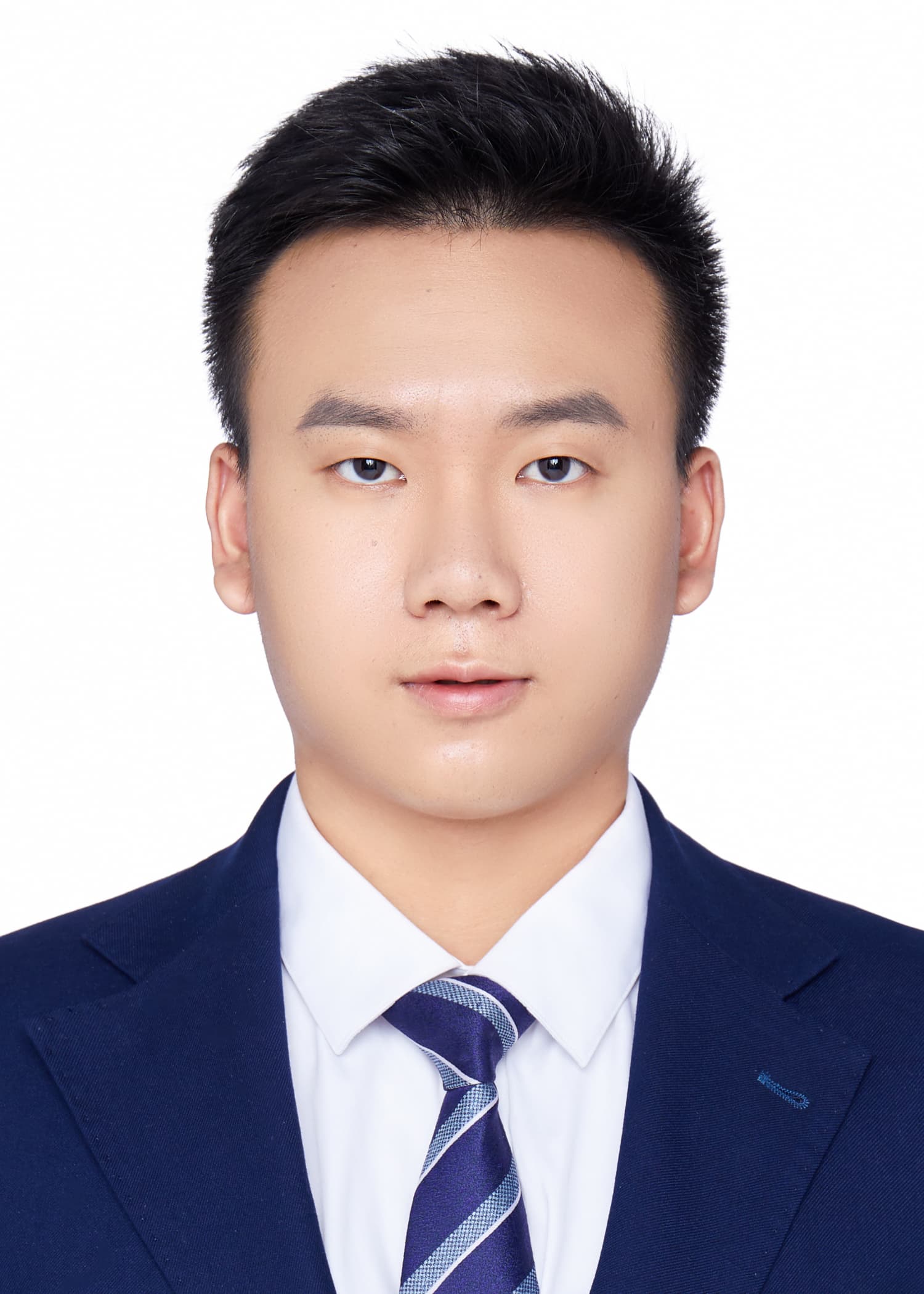}}]{Jianhua Pei} (Student Member, IEEE) received the B.Eng. degree in electrical engineering from Huazhong University of Science and Technology (HUST), Wuhan, China, in 2019. He is currently pursuing his Ph.D. degree in electrical engineering at HUST. He is also a visiting Ph.D. student with the Department of Electrical Engineering and Computer Science, Lassonde School of Engineering, York University, Canada, in 2024. His research interests include power system data quality improvement, power system cybersecurity, and artificial intelligence applications for communications.
\end{IEEEbiography}

\begin{IEEEbiography}[{\includegraphics[width=1in,height=1.25in,clip,keepaspectratio]{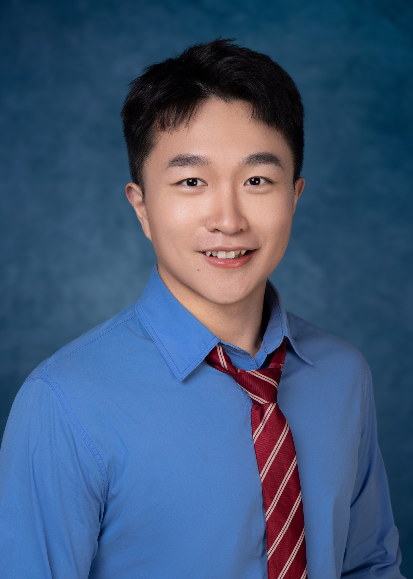}}]{Cheng Feng} (Member, IEEE) now an Ezra Postdoctoral Associate in Cornell University. He received the B.S. degree in Electrical Engineering in Huazhong University of Science and Technology in June, 2019, and the Ph.D. degree in Electrical Engineering from Tsinghua University in June, 2024. During February 2023 to August 2023, he was a visiting scholar in Automatic Control Lab (ifA), ETH Zurich. His research interests include cyber-physical system optimization and control in energy systems.
\end{IEEEbiography}

\begin{IEEEbiography}[{\includegraphics[width=1in,height=1.25in,clip,keepaspectratio]{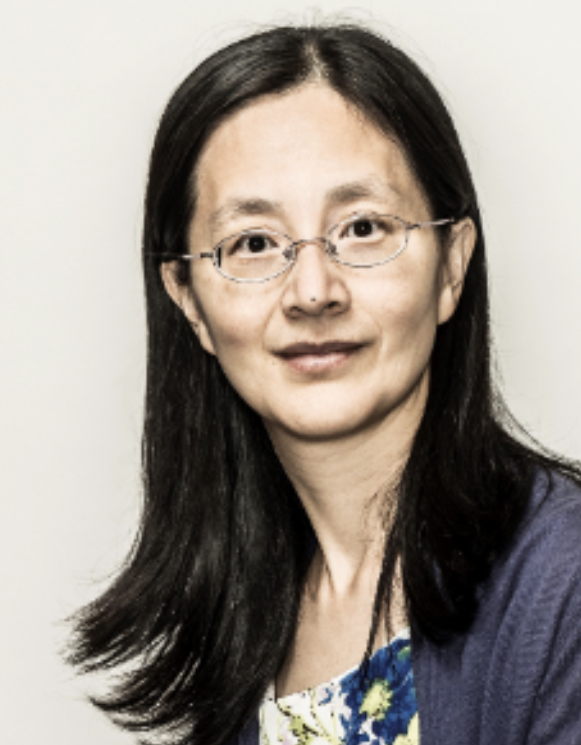}}]{Ping Wang} (Fellow, IEEE) is a Professor at the Department of Electrical Engineering and Computer Science, York University, and a Tier 2 York Research Chair. Prior to that, she was with Nanyang Technological University, Singapore, from 2008 to 2018. Her recent research interests focus on integrating Artificial Intelligence (AI) techniques into communications networks. Her scholarly works have been widely disseminated through top-ranked IEEE journals/conferences and received the IEEE Communications Society Best Survey Paper Award in 2023, and the Best Paper Awards from IEEE prestigious conference WCNC in 2012, 2020 and 2022, from IEEE Communication Society: Green Communications \& Computing Technical Committee in 2018, from IEEE flagship conference ICC in 2007. She has been serving as the associate editor-in-chief for IEEE Communications Surveys \& Tutorials and an editor for several reputed journals, including IEEE Transactions on Wireless Communications. She is
a Fellow of the IEEE and a Distinguished Lecturer of the IEEE Vehicular Technology Society (VTS). She is also the Chair of the Education Committee of IEEE VTS.
\end{IEEEbiography}

\begin{IEEEbiography}[{\includegraphics[width=1in,height=1.25in,clip,keepaspectratio]{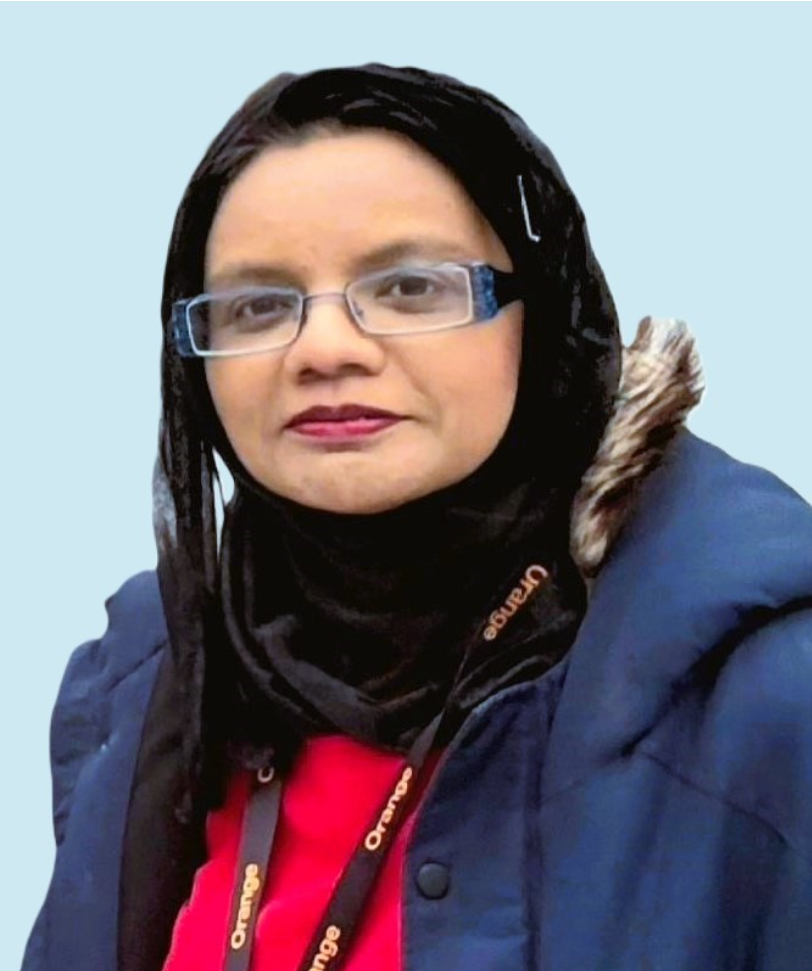}}]{Hina Tabassum} (Senior Member, IEEE) received the Ph.D. degree from the King Abdullah University of Science and Technology (KAUST). She is currently an Associate Professor with the Lassonde School of Engineering, York University, Canada, where she joined as an Assistant Professor, in 2018. She is also appointed as a Visiting Faculty at University of Toronto in 2024 and the York Research Chair of 5G/6G-enabled mobility and sensing applications in 2023, for five years. Prior to that, she was a postdoctoral research associate at University of Manitoba, Canada. She has been selected as IEEE ComSoc Distinguished Lecturer (2025-2026). She is listed in the Stanford's list of the World's Top Two-Percent Researchers in 2021-2024. She received the Lassonde Innovation Early-Career Researcher Award in 2023 and the N2Women: Rising Stars in Computer Networking and Communications in 2022. She has been recognized as an Exemplary Editor by the IEEE Communications Letters (2020), IEEE Open Journal of the Communications Society (IEEE OJCOMS) (2023-2024), and IEEE Transactions on Green Communications and Networking (2023). She was recognized as an Exemplary Reviewer (Top 2\% of all reviewers) by IEEE Transactions on Communications in 2015, 2016, 2017, 2019, and 2020. She is the Founding Chair of the Special Interest Group on THz communications in IEEE Communications Society (ComSoc)-Radio Communications Committee (RCC). She served as an Associate Editor for IEEE Communications Letters (2019-2023), IEEE OJCOMS (2019-2023), and IEEE Transactions on Green Communications and Networking (2020-2023). Currently, she is also serving as an Area Editor for IEEE OJCOMS and an Associate Editor for IEEE Transactions on Communications, IEEE Transactions on Wireless Communications, and IEEE Communications Surveys \& Tutorials.
\end{IEEEbiography}

\begin{IEEEbiography}[{\includegraphics[width=1in,height=1.25in,clip,keepaspectratio]{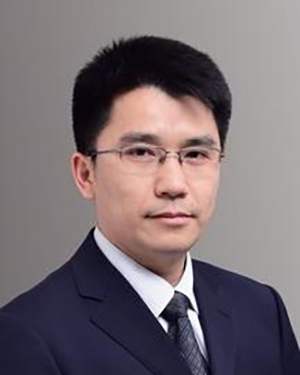}}]{Dongyuan Shi} (Senior Member, IEEE) received the B.S. and Ph.D. degrees in electrical engineering from Huazhong University of Science and Technology (HUST), China, in 1996 and 2002, respectively. From 2007 to 2009, he was a Visiting Scholar with Cornell University, Ithaca, NY. He is currently a professor with the School of Electrical and Electronic Engineering, HUST. His research interests include power system analysis and computation, cybersecurity, and software technology.
\end{IEEEbiography}

\end{document}